\documentclass[11pt,letterpaper]{article}
\usepackage{labreport}
\usepackage{float}
\usepackage{placeins}

\usepackage{amsmath,amsfonts,bm}

\def\eqref#1{equation~\ref{#1}}

\def\1{\bm{1}}

\DeclareMathAlphabet{\mathsfit}{\encodingdefault}{\sfdefault}{m}{sl}
\SetMathAlphabet{\mathsfit}{bold}{\encodingdefault}{\sfdefault}{bx}{n}

\reportlab{}
\reporttype{Technical Report}
\reportnumber{}
\reportdate{October 2026}
\reportshorttitle{PASA-KV: Attacking Action-Facing Visual Routes in World Action Models}
\reportcolors{183153}{167D8D}
\reportlinkcolor{2455A4}

\newcommand{\pasakv}{\textsc{PASA-KV}}
\setcitestyle{authoryear,round,citesep={;},aysep={,},yysep={;}}
\title{PASA-KV: Attacking Action-Facing\\Visual Routes in World Action Models}
\author{Zhongbo Wu\textsuperscript{1,2,3}, Jiakai Huang\textsuperscript{1,3,*},
Zheng Zhang\textsuperscript{3,4}, Siyu Xu\textsuperscript{5},\\[3pt]
Kaixiang Chen\textsuperscript{1,*}, Xiaohong Liu\textsuperscript{1,2},
Shan You\textsuperscript{3}, Chang Xu\textsuperscript{5}, Tao Huang\textsuperscript{1,\textdagger}}
\reportaffiliations{\textsuperscript{1}Shanghai Jiao Tong University\quad
\textsuperscript{2}Shanghai Innovation Institute\\
\textsuperscript{3}ACE Robotics\quad\textsuperscript{4}Nanyang Technological University\\
\textsuperscript{5}School of Computer Science, The University of Sydney}
\reportauthornote{\textsuperscript{*}Research intern.\quad\textsuperscript{\textdagger}Corresponding author.}
\reportresourcegroup{Project Page}{%
  \reportresourceitem{Website}{https://awesome-airsome.github.io/PASA-KV-ProjectPage/}{awesome-airsome.github.io/PASA-KV-ProjectPage/}}
\begin{document}
\maketitle
\begin{abstract}
World action models (WAMs) increasingly couple internal visual or world representations with action generation, making their world-to-action pathways a critical yet poorly understood source of robustness. We investigate when perturbations to these internal representations actually become behaviorally consequential, and uncover a pronounced mismatch between representation perturbability and closed-loop vulnerability: attack effectiveness varies sharply across visual K/V routes, while larger representation displacement does not reliably imply stronger control failure. Our analysis attributes this mismatch to the action-conditioned readout, where keys determine where action queries read and values determine what content they receive. This motivates an action-facing view of vulnerability, in which the effect of a perturbation depends on how it is routed into action generation rather than on representation-space magnitude alone. We instantiate this view with Tail-AV, a short-probe signal for identifying vulnerable routes, and operationalize it in PASA-KV, which selects action-facing K/V targets before training a fresh task-shared attack. Experiments across multiple WAM systems and a physical robot validate readout-informed target selection, demonstrate strong closed-loop attacks, and reveal substantial system-level differences in resistance.

\end{abstract}
\reportcontents
\section{Introduction}
\label{sec:introduction}
\suppressfloats[t]

World action models (WAMs) are emerging as a promising paradigm for robot control by coupling visual world representations and dynamics with action generation~\citep{ha2018worldmodels,hafner2019planet,li2026lingbotva,kim2026cosmospolicy,yuan2026fastwam}; latent-imagination methods provide an earlier route from learned world models to control~\citep{hafner2020dreamer,hafner2021dreamerv2}. Generalist robot policies include RT-1, RT-2, and Octo~\citep{brohan2023rt1,brohan2023rt2,ghosh2024octo}, while MimicGen scales demonstration generation for robot learning~\citep{mandlekar2023mimicgen}. Across direct, cascaded, and joint WAM architectures, action generation ultimately depends on internal visual or world representations. This coupling introduces a security question that cannot be answered by studying input perturbations alone: \emph{when does a change in an internal visual representation actually propagate to the action generator and alter closed-loop behavior?} Existing attacks on robot policies expose vulnerabilities at the input, representation, attention, denoising, or action level~\citep{chen2024diffusionpolicyattacker,wang2025adversarialvla,lu2026robotspatch,tae2026drift,zhang2026sarf}, but they do not reveal which world-to-action routes make WAMs behaviorally vulnerable.

\begin{figure}[t]
    \centering
    \includegraphics[width=\textwidth]{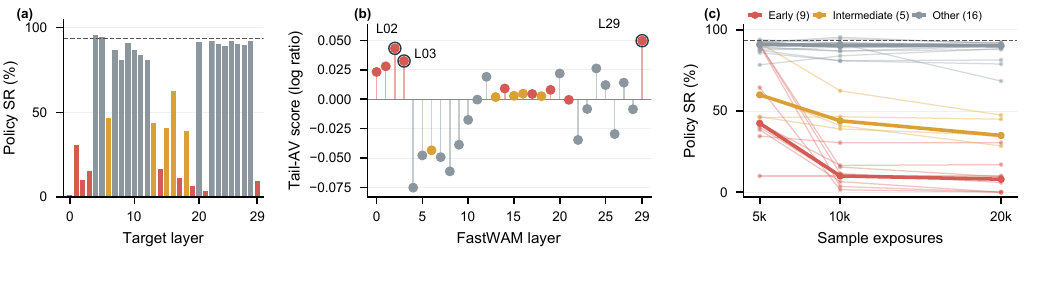}
    
    \caption{\textbf{Layerwise evidence for action-facing vulnerability.} All panels use the evaluated FastWAM checkpoint. \textbf{(a)} Policy SR for 30 singleton joint K/V targets after 10k exposures; dashed: clean reference. \textbf{(b)} Tail-AV scores from short probes; rings mark Top-3. \textbf{(c)} Budget--SR trajectories at 5k/10k/20k exposures. Colors denote retrospective response groups, unused by the selector.}
    \label{fig:intro_two_insights}
\end{figure}

A natural view is that routes whose representations can be perturbed more strongly should provide more effective attack targets. Our layerwise study shows that this view is incomplete. Under the same attack objective and training budget, perturbing different visual K/V layers of FastWAM leads to sharply different closed-loop outcomes (Figure~\ref{fig:intro_two_insights}(a,c)). More strikingly, stronger or broader internal perturbation does not necessarily produce stronger behavioral failure. Jointly attacking K and V broadens the set of effective targets, yet does not dominate either route at every layer; at L20, for example, K-only reduces policy success to 0.0\%, whereas joint K/V leaves it at 91.5\%. Across layers, the magnitude of K/V displacement likewise fails to determine eventual attack effectiveness. These results reveal a fundamental mismatch: \textbf{representation disruption is not action disruption}.

The missing information lies in how a perturbation interacts with the \emph{action-conditioned readout}. In attention-based WAMs, visual keys determine where action queries read, whereas visual values determine what content they receive. A large K displacement can therefore have little effect when it poorly aligns with the directions to which action queries are sensitive, while a large V displacement may remain inconsequential when the altered content receives little attention. K- and V-induced changes can also reinforce or partially oppose one another, providing a mechanism for the non-monotonic behavior above. This suggests that vulnerability should not be characterized solely in representation space, but through the interaction between an internal perturbation and the action-facing pathway that consumes it. We refer to this view as \emph{action-facing vulnerability}: the behavioral relevance of a perturbation depends not only on how much the representation changes, but also on how that change is routed into action generation.

This perspective changes how vulnerable routes should be identified. Rather than ranking layers by how far their representations move, we measure how a short perturbation changes the visual contribution received by action queries. This motivates Tail-AV, which combines action-query routing with the magnitude of visual content being read and, from short probes, identifies layers that later become strong attack targets (Figure~\ref{fig:intro_two_insights}(b)). Importantly, this signal is computed without closed-loop rollout outcomes. The resulting characterization naturally leads to the Probe-and-Select Action-facing K/V Attack (\pasakv{}): probe candidate routes cheaply, select those with the strongest action-facing responses, and train the final task-shared patch only on the selected routes. In this way, \pasakv{} operationalizes the action-facing view without exhaustively training and evaluating full attacks at every candidate layer.

Our experiments support this interpretation at both the mechanism and system levels. Under matched FastWAM controls with the same final K/V objective, target count, geometry, and training budget, Tail-AV-selected layers achieve 1.5\% policy success, compared with 47.5\% when layers are selected directly by probe K/V displacement and 18.5--86.5\% for evenly spaced or random selections (Section~\ref{sec:selector_validation}). The resulting \pasakv{} Top-5 attack reduces FastWAM success from 92.4\% to 0.0\%. Yet the same construction produces substantially smaller degradation on LingBot-VA and Cosmos Policy, reaching 87.2\% and 89.2\% SR, respectively, exposing pronounced architecture-dependent resistance rather than a universal vulnerability (Table~\ref{tab:main_attack_comparison}). Finally, rerunning the complete procedure on a separately trained Piper FastWAM checkpoint yields a task-shared Top-5 patch with 0/10 successes on each of the three evaluated tasks with nonzero clean success (Section~\ref{sec:real_robot}).

Our contributions are:
\begin{itemize}
    \item We characterize visual K/V vulnerability in WAMs and reveal a systematic mismatch between representation perturbability and closed-loop behavioral vulnerability: attack effectiveness varies sharply across layers, and stronger internal perturbation does not necessarily imply stronger control failure.
    \item We identify action-facing readout as the missing link between internal perturbation and behavioral consequence. This view explains the distinct roles of K and V and motivates Tail-AV as a short-probe signal for locating vulnerable visual routes.
    \item We operationalize this understanding through \pasakv{}, whose readout-informed selection substantially outperforms matched representation-based and uninformed controls. Cross-architecture and real-robot evaluations further reveal large differences in how WAMs expose and resist action-facing attacks.
\end{itemize}

\section{Related Work}
\label{sec:related_work}

\paragraph{World Action Models and Their Architectural Paradigms}
WAMs differ in how visual dynamics modeling enters action generation at
inference time~\citep{ye2026world}. Direct-policy Fast-WAM encodes observations
into visual K/V that action denoising reads without generating future
video~\citep{yuan2026fastwam}. In cascaded LingBot-VA, visual-state prediction
precedes action denoising, which accesses video-side representations through
cached K/V~\citep{li2026lingbotva}. Joint Cosmos Policy instead represents
actions and future states as latent slots within a shared denoising process,
so visual--action interactions occur within the joint transformer rather than
between sequential prediction stages~\citep{kim2026cosmospolicy}.

\paragraph{Adversarial manipulation of visual attention and representations}
Localized token and patch attacks can disrupt ViTs~\citep{joshi2021adversarialtokens,fu2022patchfool}, while patch-aware optimization improves attack transfer~\citep{wei2022transferablevit}.
At the attention level, adversarial patches can concentrate many queries on
patch-controlled keys~\citep{lovisotto2022attention}.
Related work studies visual jailbreaks~\citep{qi2024visualadv}, cross-attention
attacks on customized diffusion models~\citep{xu2024perturbing},
attention backdoors~\citep{attentionhijack2022}, and attention-head
path-deviation diagnostics~\citep{pathdeviation2026}.
UPA-RFAS combines feature, attention, and semantic objectives~\citep{lu2026robotspatch};
its attention objective targets text-to-vision aggregation, whereas PASA-KV
targets visual keys and values consumed by action queries.
AGSD studies physical action-to-vision attention hijacking, and SARF mitigates
it through visual-encoder fine-tuning~\citep{zhang2026sarf}.

\paragraph{Attacks on robot policies and vision--language--action models}
OpenVLA provides a representative action-generating platform~\citep{kim2024openvla}.
DP-Attacker targets diffusion policies~\citep{chen2024diffusionpolicyattacker};
\citet{wang2025adversarialvla} study spatially grounded untargeted and targeted
trajectory objectives; DRIFT attacks the first velocity-field step of flow-matching VLAs~\citep{tae2026drift}.
Other work uses trajectory prefixes~\citep{wang2026partiallyobservable},
targets CoT reasoning~\citep{huang2026trap} or visual proprioception~\citep{fu2026vlahijack},
steers visual-token attention~\citep{prasad2026attentionsteering}, or studies
textual jailbreaks and training-time backdoors~\citep{jones2025robogcg,zhou2025badvla}.

\section{From Representation Perturbability to Action Vulnerability}
\label{sec:insights}

We use FastWAM as a controlled testbed to study how perturbations to internal visual representations become behaviorally consequential. We define \emph{action-facing visual K/V} as the visual keys and values consumed by action queries, and optimize a single RGB patch against individual layers while keeping the victim and attack protocol fixed. This controlled setting lets us separate three quantities that are easily conflated: how much an internal representation can be changed, how much that change alters the action-facing readout, and how much the resulting policy behavior degrades. We first establish that vulnerability varies sharply across internal routes, then ask what actually explains this variation.

\subsection{Behavioral vulnerability is highly route-dependent}
\label{sec:layer_responses}

Under the same joint K/V objective and training budget, different FastWAM layers exhibit sharply different closed-loop vulnerability (Figure~\ref{fig:intro_two_insights}(a,c)). For example, L03 drops from 91.0\% Policy SR at 5k exposures to 15.5\% and 9.5\% at 10k and 20k exposures, whereas L28 remains close to clean performance over the same sweep. We retrospectively summarize these responses as Early, Intermediate, and Other layers only for analysis; they are not used for attack construction or selection.

This heterogeneity rules out a simple picture in which all internal visual routes become similarly vulnerable once they are exposed to the same attack objective. It instead suggests that some routes are intrinsically more consequential for control than others. The most direct explanation would be that these routes are simply easier to perturb in representation space: if so, the layers with larger K/V displacement should also induce stronger behavioral failures. We test this possibility next.

\subsection{Representation displacement does not explain behavioral vulnerability}
\label{sec:representation_gap}

Representation-space change captures part of the attack response, but it does not explain the layerwise variation above. Across the 30 layers, normalized K and V displacement have only limited association with attack strength,
$\rho(D_K,\mathrm{AttackStrength})=0.316$ and
$\rho(D_V,\mathrm{AttackStrength})=0.528$.
Layers with larger representation changes therefore do not reliably produce larger closed-loop failures. Complementary footprint and Rescue/Transfer analyses lead to the same conclusion and are reported in Appendix~\ref{app:layer_roles}.

The failure of this representation-space diagnostic changes the question. The issue is no longer simply \emph{how much} a layer can be perturbed, but \emph{which part of that perturbation is actually exposed to action generation}. A visual representation may move substantially while the action policy barely consumes the changed content; conversely, a smaller perturbation may matter if it lies directly on the route used by action queries. This motivates us to examine the interface through which visual K/V are read by the action generator.

\subsection{Vulnerability emerges through the action-conditioned readout}
\label{sec:kv_mechanism}
\label{sec:component_ablation}

Consider how action query $i$ reads visual token $j$:
\begin{equation}
    A_{ij}=\operatorname{softmax}_{j}\!\left(\frac{Q_iK_j^{\top}}{\sqrt d}\right),
    \qquad
    y_i=\sum_j A_{ij}V_j.
    \label{eq:action_readout}
\end{equation}
Keys determine \emph{where} action queries read, whereas values determine \emph{what} content they receive~\citep{vaswani2017attention}. Prior work questions attention weights as standalone explanations~\citep{jain2019attention}, while relevance-propagation methods extend analysis beyond attention visualization~\citep{chefer2021transformer}; causal mediation and model editing provide complementary probes of internal computations~\citep{vig2020causalmediation,meng2022rome}. Representation displacement therefore becomes behaviorally relevant only through its interaction with this action-conditioned readout.

For $q_i=Q_{\rm action}$, a first-order expansion under $k_j\mapsto k_j+\delta k_j$ and $v_j\mapsto v_j+\delta v_j$ gives
{\footnotesize
\begin{equation}
\delta y_i\approx
\underbrace{\frac{1}{\sqrt d}\sum_{j\in\mathcal V}a_{ij}(q_i^\top\delta k_j)(v_j-y_i)}_{u_K}
+
\underbrace{\sum_{j\in\mathcal V}a_{ij}\delta v_j}_{u_V}.
\label{eq:first_order_kv}
\end{equation}
}
The two terms reveal why raw displacement alone is insufficient. A large K perturbation may induce little readout change when it poorly aligns with the action query, i.e., when $q_i^\top\delta k_j$ is small. Likewise, a large V perturbation may contribute little when the affected content receives negligible attention. The relevant quantity is therefore not representation-space magnitude alone, but how the perturbation is gated by action-query routing.

This view also explains why K and V need not combine monotonically. Since
\[
\lVert u_K+u_V\rVert_2^2
=
\lVert u_K\rVert_2^2+\lVert u_V\rVert_2^2
+2\langle u_K,u_V\rangle,
\]
K- and V-induced readout changes may reinforce or partially oppose one another. The controlled component attacks exhibit this behavior. Under the same FastWAM protocol and 10k-exposure budget, joint K/V optimization yields effective attacks ($\mathrm{SR}\leq50\%$) on 13/30 layers, compared with 7/30 for K-only and 11/30 for V-only (Table~\ref{tab:fastwam_kv_complementarity}; complete pooled results appear in Appendix Table~\ref{tab:component_attack_coverage}). Joint optimization thus broadens effective-layer coverage overall, but does not dominate every layer: at L20, K-only reaches 0.0\% SR whereas joint K/V remains at 91.5\%.

Together, these results suggest a more precise account of the heterogeneity observed in Section~\ref{sec:layer_responses}. Vulnerability is governed not simply by whether an internal representation can be moved, but by how that perturbation interacts with the action-conditioned pathway that consumes it. We refer to this distinction as \emph{action-facing vulnerability}. Importantly, this is still a mechanistic account rather than a claim that readout change perfectly determines closed-loop failure. The stronger test is whether action-facing information can identify vulnerable routes \emph{before} a full attack is trained.

\subsection{Can short action-facing responses anticipate later vulnerability?}
\label{sec:influence_signals}

The action-facing view makes a concrete prediction: if the readout captures information about vulnerability beyond raw representation displacement, then an early readout-aware response should help distinguish routes that later become effective attack targets. We test this prediction with short probes rather than full attack training.

For visual token $j$, we define its action-facing contribution as
\begin{equation}
    I_{ij}=A_{ij}\lVert V_j\rVert_2,
    \label{eq:action_contribution}
\end{equation}
which combines whether the action query reads a visual token with the magnitude of the content delivered through that route. Tail-AV summarizes the probe-induced growth of the sparse upper tail of these contributions. Under the FastWAM protocol, Probe100 Tail-AV prioritizes L29/L02/L03, which later become strong attack targets under longer optimization (Figure~\ref{fig:intro_two_insights}(b)), without using rollout outcomes or retrospective response labels.

This prospective behavior is not merely a consequence of one favorable probe run. Repeated Probe100 runs produce highly stable rankings, while comparisons with representation- and component-based signals further support the utility of action-facing responses; complete statistics are reported in Appendix~\ref{app:probe_prediction} and Section~\ref{app:selector_controls}. We emphasize that Tail-AV is a short-budget ranking signal rather than a calibrated surrogate for final behavioral damage.

The remaining question is whether this ranking advantage survives the step that ultimately matters: training a full attack under a matched budget. Section~\ref{sec:selector_validation} therefore compares Tail-AV-selected routes against representation-based and uninformed selections while holding the final objective, target count, geometry, and training budget fixed. If action-facing vulnerability captures something genuinely useful beyond representation perturbability, its selected routes should support stronger final attacks. Section~\ref{sec:method} turns exactly this criterion into \pasakv{}.

\FloatBarrier
\section{From Characterization to Attack: PASA-KV}
\label{sec:method}

\begin{figure}[!htbp]
    \centering
    \begin{minipage}{\textwidth}
        \centering
        \includegraphics[width=\textwidth]{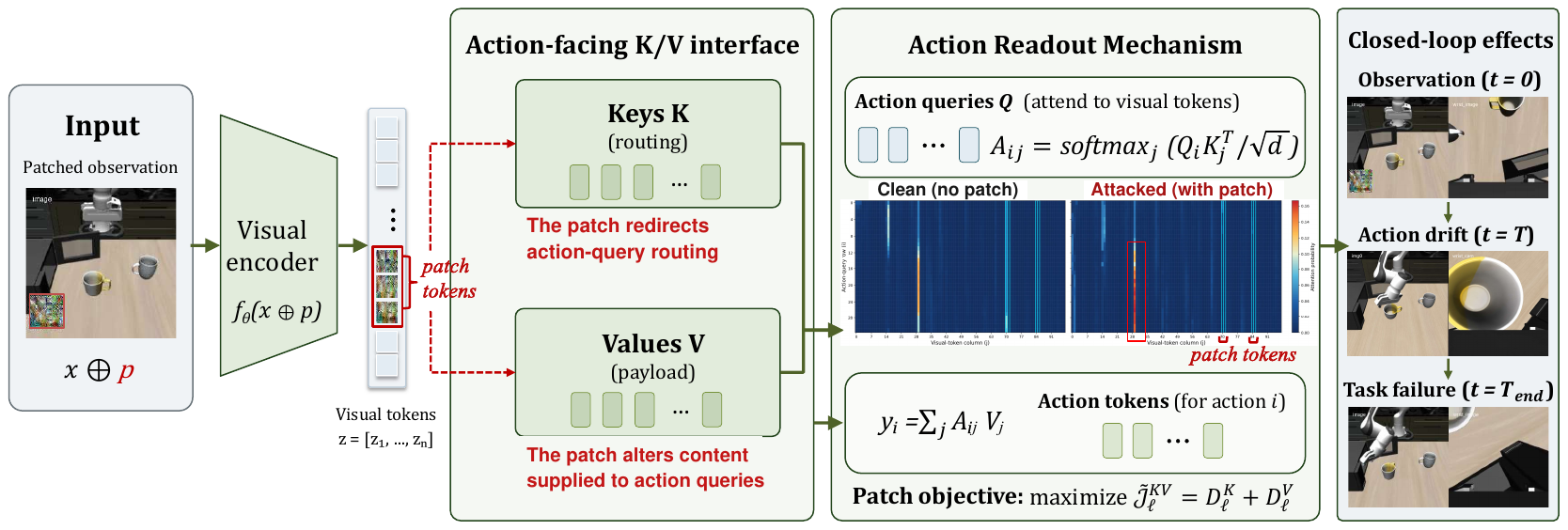}
        \par\vspace{1mm}
        {\small\textbf{(a)} PASA-KV attack surface: action-facing keys and values\par}
    \end{minipage}
    \vspace{2mm}

    \begin{minipage}{\textwidth}
        \centering
        \includegraphics[width=\textwidth]{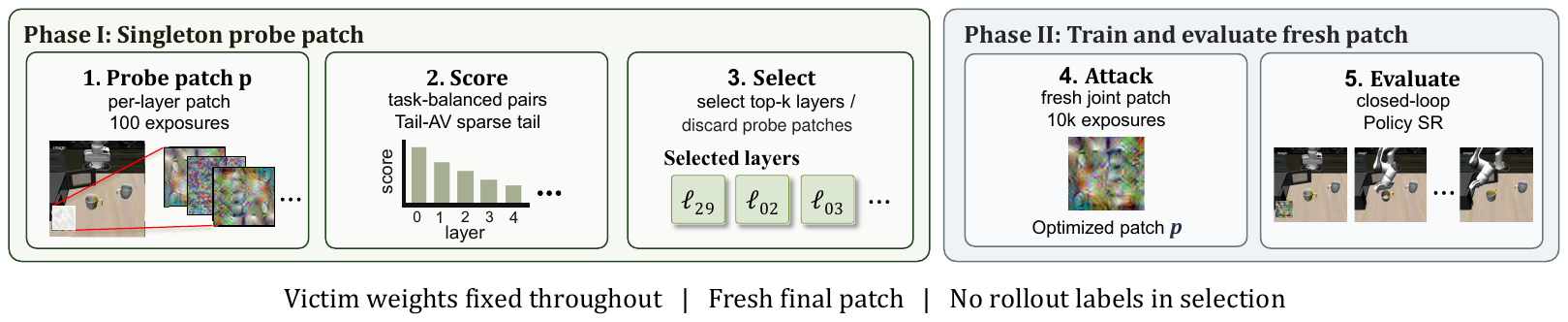}
        \par\vspace{1mm}
        {\small\textbf{(b)} PASA-KV pipeline: locate, select, and train a fresh attack\par}
    \end{minipage}
    \caption{\textbf{PASA-KV construction pipeline.}
    \textbf{(a)} Action-facing visual K/V interface.
    \textbf{(b)} Candidate routes are first probed with a small optimization budget and ranked by Tail-AV; a fresh task-shared patch is then trained only on the selected Top-$k$ routes.
    Victim weights remain fixed throughout, and rollout outcomes are excluded from layer selection.}
    \label{fig:method}
\end{figure}

Section~\ref{sec:insights} shows that internal routes differ substantially in behavioral vulnerability, while representation displacement alone provides an incomplete basis for deciding where to attack. The practical difficulty is that directly resolving this uncertainty is expensive: fully training and evaluating an attack at every candidate layer would spend most of the budget merely discovering which routes matter. \pasakv{} addresses this problem by \emph{decoupling route identification from full attack optimization}. As illustrated in Figure~\ref{fig:method}, it first uses inexpensive singleton probes to expose the short-budget action-facing response of each candidate route, selects promising layers with Tail-AV, and only then allocates the full optimization budget to a fresh multi-route attack.

\subsection{Attack setting and route probing}

We consider a known-task white-box attack~\citep{goodfellow2015adversarial,madry2018robust} with a frozen victim and ten LIBERO-10 tasks~\citep{liu2023libero}. A single RGB patch $p$ is shared across tasks, drawing on localized patch and physical-world attack formulations~\citep{brown2017adversarialpatch,athalye2018eot,eykholt2018physical}. During training, the patch modifies clip frame 0; at deployment, it is applied to the current primary observation at every replan without rewriting history. Primary WAM protocols use current-view patching without transformation sampling. Architecture-specific interfaces, geometries, and deployment details appear in Section~\ref{app:registered_threats}.

The first stage is designed only to reveal how each candidate route responds under a small, common perturbation budget. For a candidate layer $\ell$, we optimize a singleton probe patch against the relative displacement of its visual K/V:
\begin{equation}
 D^Z_\ell=
 \frac{\lVert Z_\ell(x\oplus p)-Z_\ell(x)\rVert_F}
 {\max(\lVert Z_\ell(x)\rVert_F,\epsilon)},
 \quad Z\in\{K,V\},
 \qquad
 J^{KV}_\ell=\tfrac12\mathbb{E}[D^K_\ell+D^V_\ell].
 \label{eq:probe}
\end{equation}
The target layer $\ell$ specifies the K/V loss location; the resulting patch can also affect representations at other layers. FastWAM uses $\epsilon=10^{-8}$ and 100 exposures per candidate layer (Probe100). Only RGB patch pixels are optimized through the frozen victim, without unrolling action denoising~\citep{chi2023diffusionpolicy,ho2020ddpm,song2021score,lipman2023flowmatching,liu2023rectifiedflow}; optimization and capture details appear in Section~\ref{app:optimization}.

Probe patches optimize K/V displacement; Tail-AV ranks the resulting action-facing responses.

\subsubsection{Patch parameterization, optimization, and gradient path}
\label{app:optimization}

Probe100 optimizes a singleton K/V loss for 100 exposures per candidate layer. Final training resets to the designated initial patch and optimizes the mean selected-layer loss for 10,000 BS1 updates; probe pixels and optimizer state are discarded. The primary suite uses protocol/run label $42$, while singleton configurations use seed $20260801$. Thus, $L$ candidates require $100L+10{,}000$ exposures plus scoring forwards. Table~\ref{tab:fastwam_attack_recipe} gives the optimizer.

\begin{table}[htbp]
\centering
\caption{FastWAM PASA-KV attack-construction recipe. These settings optimize RGB patch pixels with the victim weights frozen. Probe and final training share the optimizer; only the loss-target set and update budget change.}
\label{tab:fastwam_attack_recipe}
\footnotesize
\begin{tabular}{@{}p{0.26\linewidth}p{0.69\linewidth}@{}}
\toprule
Setting & Documented implementation \\
\midrule
Trainable variable & FP32 RGB patch, $3\times50\times50$; frozen victim, BF16 forward \\
Optimizer & AdamW; learning rate $5\times10^{-4}$, betas $(0.9,0.95)$, epsilon $10^{-8}$, weight decay $0.01$ \\
Schedule and gradients & Constant learning rate; no warm-up, scheduler, gradient clipping or accumulation in this patch loop \\
Initialization & Uniform $[0,1]$ initializer; primary attack suite indexed by protocol/run label $42$; singleton training configurations record seed $20260801$; final training starts from fresh pixels \\
Projection and loss & Clamp pixels to $[0,1]$ after every update; minimize $-J^{KV}$ (singleton) or $-J_{\rm joint}$; image-L1 weight zero \\
Updates and batch & BS1, one update per exposure; 100 per singleton probe, 10,000 per final patch \\
Training inputs & LIBERO-10 observation clips; documented scoring-input exclusion list; clip frame 0 primary view only is patched \\
Patch selection & Update 100 for probes; update 10,000 for final evaluation; no rollout-based checkpoint selection \\
\bottomrule
\end{tabular}
\end{table}

\paragraph{Patch composition and strength}
With RGB input $x\in[0,1]$, binary rectangle mask $M$, and embedded patch $P(p)$, $x\oplus p=(1-M)\odot x+M\odot P(p)$. Patch values are converted to $[-1,1]$ before replacement in the normalized video tensor; all other pixels and frames are unchanged. This recipe has no blending, transformation sampling, or pixel-distance penalty. Tail-AV scores each update-100 probe at full strength against the unpatched input.

\paragraph{Loss tensors and gradient path}
FastWAM encodes fixed clips at video timestep zero. Equation~\ref{eq:probe} computes FP32 relative visual-cache norms (denominator floor $10^{-8}$) and averages K/V components, selected layers, and batch samples equally. The clean cache has no gradients; the attacked path backpropagates only to the FP32 patch, with the victim frozen. Runs reset their designated initial pixels and optimizer states, clamp pixels after every update, and evaluate update 10,000. Headline Top-3/4/5 patches use Eval500 (trials 0--49 per task); evaluation seeds are 42 for FastWAM/LingBot and 195 for Cosmos extensions.

\subsection{Action-facing route selection}
\label{sec:short_probe_selection}

Following Equation~\ref{eq:action_contribution}, we characterize each probed layer through the contribution
$I=A\lVert V\rVert_2$, which combines action-query routing with the magnitude of visual content being read. For task $t$ and candidate layer $\ell$, we flatten $I$ over $H$ heads, $Q$ action queries, and $T$ visual tokens. Let $\mu_q(I)$ denote the mean of its largest $\lceil qHQT\rceil$ entries, with $q=0.005$. Tail-AV measures the relative growth of this sparse upper tail between the patched and clean forwards:
\begin{equation}
 s_\ell^{\mathrm{Tail}}
 =
 \frac{1}{10}\sum_{t=1}^{10}
 \log\!\left[
 \max\!\left(
 \frac{\mu_q(I^p_{\ell t})}
 {\max(\mu_q(I^0_{\ell t}),\delta)},
 \delta
 \right)
 \right],
 \qquad
 \delta=10^{-12}.
 \label{eq:tail_av_score}
\end{equation}

The score averages the log ratio from one held-out scoring observation per task across the ten known tasks. These observations are excluded from patch optimization, and no closed-loop rollout outcome enters the score. Candidate layers are ranked by descending $s_\ell^{\mathrm{Tail}}$, with ascending layer ID as the tie-breaker, and the Top-$k$ set $S$ is retained. Scoring and capture details appear in Section~\ref{app:selectors}; complete layerwise rankings are retained in Appendix Table~\ref{tab:selector_full_rankings}.

\subsubsection{Scoring forwards and architecture-specific capture interfaces}
\label{app:selectors}

Tail-AV uses ten fixed clean/patched scoring pairs, one per task, held out from patch optimization. Scoring forwards provide internal responses and simulator rollouts provide Policy SR. Equation~\ref{eq:tail_av_score} takes the largest $\lceil0.005HQT\rceil$ clean and attacked influences independently, forms their floored log ratio, and averages tasks equally.

FastWAM scores fixed-clip visual caches at the first action-denoising step. After masked softmax over all keys, it selects current-frame visual columns without within-span renormalization. Influence has shape $[1,H,Q,T]$; clean/patched forwards share action noise. Full-strength FastWAM probes require $30\times10=300$ scoring pairs.

\paragraph{Architecture-specific interfaces}
Table~\ref{tab:selector_implementation_scope} specifies each capture interface. LingBot-VA and Cosmos Policy use the same top-0.5\% task-mean log ratio over $[1,H,Q,T]$ influence, with 30 and 28 candidate layers, respectively.

\begin{table}[t]
\centering
\caption{Architecture-specific selector capture interfaces. The common influence reduction is specified in Section~\ref{app:selectors}.}
\label{tab:selector_implementation_scope}
\footnotesize
\begin{tabular}{@{}>{\raggedright\arraybackslash}p{0.16\linewidth}>{\raggedright\arraybackslash}p{0.37\linewidth}>{\raggedright\arraybackslash}p{0.39\linewidth}@{}}
\toprule
System & Scoring interface & Attention normalization and visual slice \\
\midrule
FastWAM & Fixed-clip cache; first action-denoising step; 30 layers & Masked softmax over available keys, then current-frame visual columns; no within-slice renormalization. \\
LingBot-VA & One video step (endpoint 0.6), followed by the first of ten action steps; 30 layers & Action queries attend to full video-cache and action keys; retain current primary-view cache columns and values, conditional batch only. \\
Cosmos Policy & One joint-noise level; action-slot 4 queries; blocks 0--27 & Softmax over all joint-slot keys; retain current-primary slot 3 columns and values. Slot indices are zero-based. \\
\bottomrule
\end{tabular}
\par\smallskip
\begin{minipage}{\linewidth}
\footnotesize
Each interface forms $I=A\lVert V\rVert_2$ without renormalizing attention inside the selected visual slice. The selected layer sets are listed in the appendix text; Top-$k$ outcomes, denominators, and recorded construction costs appear in main-text Table~\ref{tab:main_attack_comparison}.
\end{minipage}
\end{table}

\paragraph{Selection protocol}
Rules and layer sets use the known LIBERO-10 task suite and are fixed before final-patch rollouts. Each Top-$k$ prefix in Table~\ref{tab:selector_full_rankings} is used to train a fresh joint patch.

\subsection{Fresh multi-route attack training}

After selection, we deliberately discard all singleton probe patches and their optimizer states. A fresh task-shared patch is initialized and jointly optimized over the selected routes:
\begin{equation}
    J_{\mathrm{joint}}(p;S)
    =
    |S|^{-1}\sum_{\ell\in S}J^{KV}_\ell(p).
    \label{eq:joint_final}
\end{equation}
The final patch is trained for 10k exposures with the victim fixed throughout. For $L$ candidate layers, the complete construction therefore requires $100L$ probe exposures followed by one 10k-exposure final optimization, in addition to the lightweight scoring forwards.

Resetting the patch ensures that Tail-AV contributes only the selected layer set $S$; all matched selector-control attacks learn their perturbation anew under the same full-training objective and budget.

The resulting procedure is therefore intentionally simple: Probe100 asks how each candidate route responds under a common small budget, Tail-AV identifies which responses are most exposed to action generation, and fresh joint training concentrates the full budget on those routes.

\section{Experimental Setup}
\label{sec:protocol}
We evaluate direct FastWAM, cascaded LingBot-VA, and joint Cosmos Policy on LIBERO-10 under their documented inference settings. FastWAM uses our frozen checkpoint and released inference configuration; LingBot-VA and Cosmos Policy use reduced-step solvers.

\subsection{Threat Models}
\label{app:registered_threats}

All attacks use frozen victims on known tasks, with checkpoint selection preceding patch training. Table~\ref{tab:experiment_map} records patch lineage and evaluation denominators.

The exact checkpoint provenance is recorded in Table~\ref{tab:fastwam_victim_config}. The frozen FastWAM victim uses precomputed text embeddings and FP32 training; attacks use BF16 inference.

\begin{table}[h]
\centering
\caption{Architecture-specific main threat models. Historical frames and cached states are not retroactively modified.}
\label{tab:registered_threats}
\footnotesize
\setlength{\tabcolsep}{2pt}
\begin{tabular}{@{}lp{0.30\textwidth}p{0.13\textwidth}p{0.20\textwidth}r@{}}
\toprule
Victim & Inference graph and visual route & Patch (px @ $x,y$) & Evaluation solver & Clean \\
\midrule
FastWAM & Direct; current-primary action-facing video K/V & $50\times50$ at $(5,160)$ & 10 action steps & 462/500 (92.4\%) \\
LingBot-VA & Cascaded; current agent-view video-to-action route & $29\times29$ at $(3,91)$ & 1 video + 10 action steps & 462/500 (92.4\%) \\
Cosmos Policy & Joint; current-primary conditioning slot & $50\times50$ at $(5,160)$ & 1 joint denoising step & 484/500 (96.8\%) \\
\bottomrule
\end{tabular}
\end{table}

\begin{figure}[t]
\centering
\includegraphics[width=\textwidth]{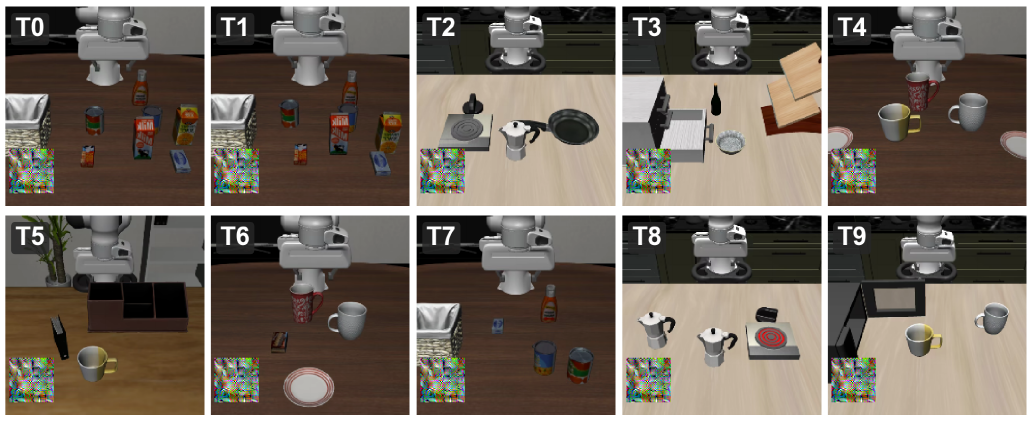}
\caption{\textbf{FastWAM patch geometry on LIBERO-10.} Tasks T0--T9 share $(x,y,w,h)=(5,160,50,50)$ on the $224\times224$ primary view (4.98\% of that view; 2.49\% of the two-view input). The secondary view is unpatched.}
\label{fig:fastwam_patch_positions}
\end{figure}

\paragraph{Inference and deployment}
FastWAM encodes video once at time 0 before ten action steps. Main patches lie outside manipulation regions, patch clip frame 0 during training, and patch each current observation at deployment. Table~\ref{tab:registered_threats} gives Eval500 references; reduced-step controls and alternate geometries are documented separately below.

\subsection{Reduced-Step Clean-Performance Controls}
\label{app:reduced_step_controls}

LingBot/Cosmos use reduced-step solvers: one video step (endpoint $s=0.6$) plus ten action steps, and one joint step, respectively. FastWAM retains its released configuration (Section~\ref{app:registered_threats}). Table~\ref{tab:reduced_step_controls} reports clean performance under these settings.

\begin{table}[h]
\centering
\footnotesize
\caption{\textbf{Reduced-step clean controls.} Each condition has 200 episodes; bracketed values are two-sided Wilson 95\% intervals. Drops use the high-step reference.}
\label{tab:reduced_step_controls}
\setlength{\tabcolsep}{3pt}
\begin{tabular}{@{}l c l r l r@{}}
\toprule
System & Solver & Use & Success & Clean SR [95\% CI] & Drop \\
\midrule
LingBot-VA & v20/a50 & High ref. & 199/200 & 99.5\% [97.22, 99.91] & --- \\
LingBot-VA & v3/a10 & Mid. & 191/200 & 95.5\% [91.67, 97.61] & $-4.0$ pp \\
LingBot-VA & v1/a10 & Unified & 187/200 & 93.5\% [89.20, 96.16] & $-6.0$ pp \\
\midrule
Cosmos & j5 & High ref. & 198/200 & 99.0\% [96.43, 99.73] & --- \\
Cosmos & j3 & Mid. & 191/200 & 95.5\% [91.67, 97.61] & $-3.5$ pp \\
Cosmos & j1 & Unified & 194/200 & 97.0\% [93.61, 98.62] & $-2.0$ pp \\
\bottomrule
\end{tabular}
\end{table}

The selected settings retain 93.5\%/97.0\% clean SR, 6.0/2.0 points below their high-step references.

\subsection{Baseline Adaptation and Evaluation}
External baselines use the method-specific geometries and observation pools listed in Appendix~\ref{app:baselines}. Section~\ref{sec:selector_validation} isolates route selection using controls with a shared final K/V objective, target count, geometry, and training budget. Appendix~\ref{app:baselines} records each baseline's fidelity to its source principle. The primary attack comparisons use Eval500; matched selector controls use a common Eval200 grid. These evaluation denominators and patch lineages are kept separate in Table~\ref{tab:experiment_map}.

\section{Attack Results and Selector Validation}
\label{sec:experiments}

\subsection{Main results: closed-loop attacks and construction costs}
\label{sec:main_results}

Table~\ref{tab:main_attack_comparison} shows that Top-5 reaches the lowest observed SR on FastWAM, LingBot-VA, and Cosmos (0.0/87.2/89.2\%; tied on Cosmos). Recorded PASA-KV construction costs are lower than the listed adaptations on FastWAM and LingBot-VA; on Cosmos, DP-Attacker's 2.88 GPU-h is marginally lower than PASA-KV Top-3's 2.98 GPU-h. The smaller effects on LingBot-VA and Cosmos Policy highlight these system-level differences.

\begin{table}[!htbp]
    \centering
    \caption{\textbf{Fixed-patch comparisons with WAM-adapted baselines.} Policy SR (\%, lower is stronger). All conditions use LIBERO-10 Eval500 (10 tasks $\times$ 50 trials). Each PASA-KV Top-$k$ condition uses a separately trained, task-shared patch (10k exposures); bold marks the lowest observed SR per architecture. GPU-h reports recorded construction cost; PASA-KV values include Probe100, selection, and final training for all three WAMs (Section~\ref{app:costs}).}
    \label{tab:main_attack_comparison}
    \footnotesize
    \setlength{\tabcolsep}{1pt}
    \begin{tabular}{@{}lcccccc l@{}}
        \toprule
        & \multicolumn{2}{c}{FastWAM} & \multicolumn{2}{c}{LingBot-VA} & \multicolumn{2}{c}{Cosmos Policy} & \\
        \cmidrule(lr){2-3}\cmidrule(lr){4-5}\cmidrule(lr){6-7}
        Method & SR $\downarrow$ & GPU-h $\downarrow$ & SR $\downarrow$ & GPU-h $\downarrow$ & SR $\downarrow$ & GPU-h $\downarrow$ & Protocol \\
        \midrule
        Shared clean reference & 92.4 & -- & 92.4 & -- & 96.8 & -- & Shared Eval500 reference \\
        \midrule
        DP-Attacker & 81.0 & 4.23 & 94.2 & 1.97 & 95.2 & 2.88 & Diffusion \\
        DRIFT & 53.0 & 4.78 & 92.4 & 2.46 & \textbf{89.2} & 3.33 & Flow/EDM \\
        UPA-RFAS & 90.8 & 10.75 & 91.0 & 8.94 & \textbf{89.2} & 23.74 & No PSM \\
        UPA & 77.4 & 6.45 & 94.2 & 8.65 & 89.6 & 3.14 & Action \\
        TMA & 81.2 & 6.12 & 93.0 & 8.49 & 90.2 & 3.13 & Action \\
        UADA & 71.4 & 6.25 & 95.0 & 8.64 & \textbf{89.2} & 3.09 & Action \\
        \midrule
        PASA-KV (Tail-AV Top-3) & 1.20 & 2.33 & 88.0 & 0.96 & 89.6 & 2.98 & Probe--select \\
        PASA-KV (Tail-AV Top-4) & 4.40 & 2.40 & 90.2 & 1.10 & 89.6 & 3.39 & Probe--select \\
        PASA-KV (Tail-AV Top-5) & \textbf{0.00} & 2.35 & \textbf{87.2} & 1.55 & \textbf{89.2} & 3.40 & Probe--select \\
        \bottomrule
    \end{tabular}
\end{table}

\subsection{Construction cost}
\label{app:costs}

For all three WAMs, PASA-KV costs include Probe100, selection, and final patch training; they exclude victim training, smoke tests, queueing, and rollout evaluation. FastWAM Top-3/4/5 cost 2.3297/2.3961/2.3540 GPU-h, each with the retained probe charge (Table~\ref{tab:tailav_construction_cost}). Against the largest baseline total of 10.746293 GPU-h (no-PSM UPA-RFAS), these recorded costs are 77.7--78.3\% lower.

\begin{table}[H]
\centering
\caption{FastWAM Tail-AV Top-3 construction cost (GPU-h). The total includes probing, selection, and final patch training.}
\label{tab:tailav_construction_cost}
\footnotesize
\begin{tabular}{@{}lrr@{}}
\toprule
Method & Probe & Full construction \\
\midrule
Tail-AV Top-3 & 0.568632 & 2.3297 \\
\bottomrule
\end{tabular}
\end{table}

\subsection{Matched controls validate readout-informed selection}
\label{sec:selector_validation}

\begin{figure}[!htbp]
\centering
\includegraphics[width=0.90\textwidth]{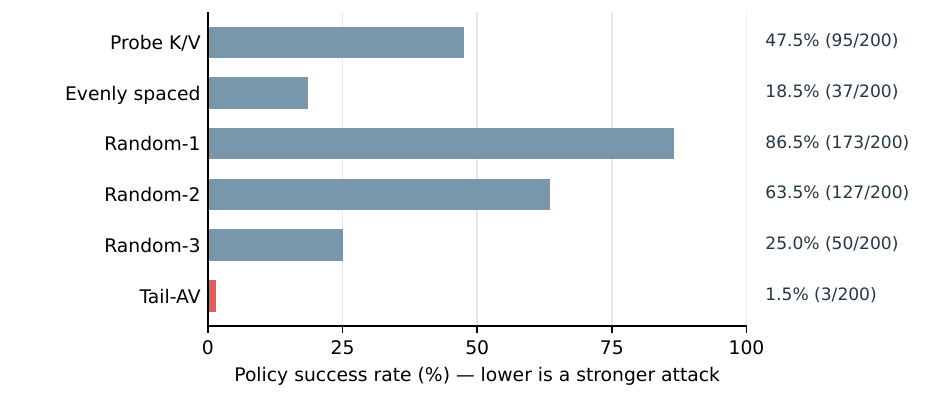}
\caption{\textbf{Matched FastWAM route-selection controls.} Bars show observed Policy SR on Eval200 (ten tasks, 20 trials each); lower is a stronger attack. Labels give successes/200. All methods share the victim, final joint K/V objective, three targets, patch geometry, and 10k exposures. The complete layer sets and retrospective classes are retained in Appendix Table~\ref{tab:selector_efficacy_controls_appendix}.}
\label{fig:selector_controls}
\end{figure}

The matched controls fix the victim, final K/V loss, target count, geometry, and 10k budget. Tail-AV obtains 1.5\% SR, compared with 47.5\% for Probe-$J^{KV}$, 18.5\% for evenly spaced selection, and 25.0--86.5\% for fixed random sets (Figure~\ref{fig:selector_controls}). The 46-point gap to Probe-$J^{KV}$ supports readout-informed selection over optimized K/V displacement alone.

\subsubsection{Matched-control protocol and repeat stability}
\label{app:selector_controls}

All six conditions in main-text Figure~\ref{fig:selector_controls} share three targets, the victim, final joint K/V loss, geometry, BS1, and 10k exposures. Controls use the common 200-episode grid (tasks 0--9; trials 0--19); deterministic rules fix random sets before training. The primary Tail-AV patch yields 3/200 successes on this grid and 6/500 on Eval500.

\paragraph{Fixed-layer final-training repeats}
Table~\ref{tab:selector_training_repeats} fixes layer sets, objective, 10k budget, exclusion manifest, and exact initial patch pixels (matching archived SHA-256). Seeds 42/43/44 change the shuffled training sample order and may affect other RNG-dependent computation; evaluation uses seed 42 on the same task/trial grid. Tail-AV has lower SR than every control in each run.

\begin{table}[H]
\centering
\caption{\textbf{Fixed-layer final-training repeats on FastWAM.} Counts are successes per 200 episodes; lower is stronger attack. Seeds 42/43/44 vary training RNG and shuffled sample order with identical saved initial pixels and layer sets. Evaluation uses seed 42 on a common Eval200 grid. Mean/SD are across runs. The seed-42 Tail-AV result uses the Eval200 subset of the primary Eval500 evaluation.}
\label{tab:selector_training_repeats}
\footnotesize
\begin{tabular}{@{}lrrrrr@{}}
\toprule
Method & Seed 42 & Seed 43 & Seed 44 & Mean SR (\%) & SD (pp) \\
\midrule
Tail-AV Top-3 & 3 & 3 & 3 & 1.50 & 0.00 \\
Probe-$J^{KV}$ & 95 & 94 & 89 & 46.33 & 1.61 \\
Evenly spaced & 37 & 53 & 43 & 22.17 & 4.04 \\
Random-1 & 173 & 170 & 169 & 85.33 & 1.04 \\
Random-2 & 127 & 131 & 11 & 44.83 & 34.08 \\
Random-3 & 50 & 19 & 25 & 15.67 & 8.22 \\
\bottomrule
\end{tabular}
\end{table}

\paragraph{Probe-run selector reproducibility}
The eight A/B/C runs are separate 30-layer Probe100 and scoring executions at recorded seeds 42--44. They share a saved probe initializer and ten scoring observations; within a seed, the scoring action-noise seed rule is identical. Across seeds, probe training and scoring action noise are both reseeded. Correlation with the canonical ranking is 0.994216--0.998665 (mean 0.996941); Top-4/5 are identical and Top-3 differs once (L01 for L03).

\begin{table}[H]
\centering
\caption{\textbf{FastWAM Tail-AV ranking repeatability on fixed scoring observations.} A/B/C are separate probe-and-score executions at one seed, sharing initial pixels and scoring inputs. Spearman correlations compare eight 30-layer rankings with the canonical ranking; Top-$k$ columns give exact set agreement. Only Seed44-B swaps L03 for L01 at Top-3.}
\label{tab:selector_rank_stability}
\footnotesize
\setlength{\tabcolsep}{6pt}
\begin{tabular}{@{}lcrrrr@{}}
\toprule
Run & Probe seed & Spearman & Top-3 & Top-4 & Top-5 \\
\midrule
Canonical & 42 & 1.000000 & Ref. & Ref. & Ref. \\
Seed42-A & 42 & 0.998665 & Yes & Yes & Yes \\
Seed42-B & 42 & 0.997775 & Yes & Yes & Yes \\
Seed43-A & 43 & 0.998665 & Yes & Yes & Yes \\
Seed43-B & 43 & 0.996885 & Yes & Yes & Yes \\
Seed43-C & 43 & 0.994216 & Yes & Yes & Yes \\
Seed44-A & 44 & 0.996440 & Yes & Yes & Yes \\
Seed44-B & 44 & 0.995996 & No & Yes & Yes \\
Seed44-C & 44 & 0.996885 & Yes & Yes & Yes \\
\midrule
Mean (8 additional) & -- & 0.996941 & 7/8 & 8/8 & 8/8 \\
\bottomrule
\end{tabular}
\end{table}

\begin{figure}[!htbp]
\centering
\includegraphics[width=0.82\textwidth]{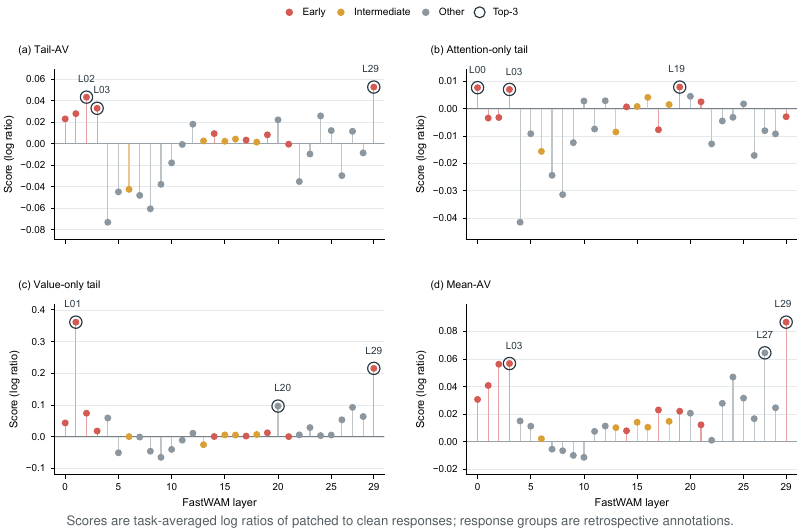}
\caption{\textbf{Complete selector-score component rankings on the evaluated FastWAM checkpoint.} Thirty-layer rankings for Tail-AV, Attention-only-tail, Value-only-tail, and Mean-AV from the shared Probe100 component-ablation run. Colors denote retrospective Early, Intermediate, and Other groups; rings mark each ranking's Top-3.}
\label{fig:component_ablation_rankings}
\end{figure}

\paragraph{Component-score rankings}
Table~\ref{tab:selector_component_ablation} and Figure~\ref{fig:component_ablation_rankings} compare all 30 layers from the shared Probe100 component-ablation run. Attention-only-tail retains routing alone; Tail-AV additionally includes the magnitude of received visual content. The two are the strongest continuous signals, while Mean-AV and Value-only-tail are weaker. Retrospectively, Tail-AV has the highest Early-group AUC (0.8624) and the strongest Early-layer coverage at Top-5 (4/5) among these score variants.

\begin{table}[H]
\centering
\caption{\textbf{Selector-score component ablation on the evaluated FastWAM checkpoint.} All four score variants use the same Probe100 component-ablation run (30 layers and ten task-balanced scoring pairs). $\rho_{10\mathrm{k}}$ and $\rho_{20\mathrm{k}}$ are Spearman correlations with later layer AttackStrength; Early-group AUC and Early@Top-$k$ are retrospective diagnostics using the nine Early layers as positives.}
\label{tab:selector_component_ablation}
\footnotesize
\setlength{\tabcolsep}{5pt}
\begin{tabular}{@{}lccccc@{}}
\toprule
Score & Early-group AUC $\uparrow$ & $\rho_{10\mathrm{k}}$ & $\rho_{20\mathrm{k}}$ & Early@Top-3 & Early@Top-5 \\
\midrule
Tail-AV & \textbf{0.8624} & \textbf{0.521} & 0.508 & \textbf{3/3} & \textbf{4/5} \\
Attention-only-tail & 0.7672 & 0.506 & \textbf{0.537} & \textbf{3/3} & 3/5 \\
Mean-AV & 0.7831 & 0.229 & 0.173 & 2/3 & 3/5 \\
Value-only-tail & 0.6931 & 0.100 & 0.143 & 2/3 & 3/5 \\
\bottomrule
\end{tabular}
\end{table}

\FloatBarrier

\begin{figure}[!htbp]
    \centering
    \begin{minipage}[c]{0.495\textwidth}
        \centering
        \setbox0=\hbox{\includegraphics[width=\linewidth]{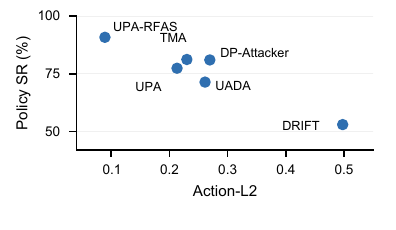}}%
        \makebox[0pt][r]{\raisebox{.5\ht0}{\footnotesize\textbf{(a)}\hspace{0.35em}}}\box0
    \end{minipage}\hfill
    \begin{minipage}[c]{0.495\textwidth}
        \centering
        \setbox0=\hbox{\includegraphics[width=\linewidth]{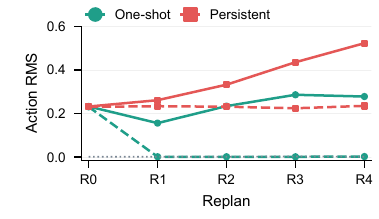}}%
        \makebox[0pt][r]{\raisebox{.5\ht0}{\footnotesize\textbf{(b)}\hspace{0.35em}}}\box0
    \end{minipage}
    \par
    \begin{minipage}[c]{0.495\textwidth}
        \centering
        \setbox0=\hbox{\includegraphics[width=\linewidth]{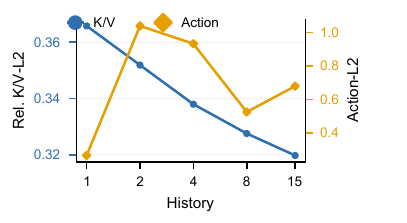}}%
        \makebox[0pt][r]{\raisebox{.5\ht0}{\footnotesize\textbf{(c)}\hspace{0.35em}}}\box0
    \end{minipage}\hfill
    \begin{minipage}[c]{0.495\textwidth}
        \centering
        \setbox0=\hbox{\includegraphics[width=\linewidth]{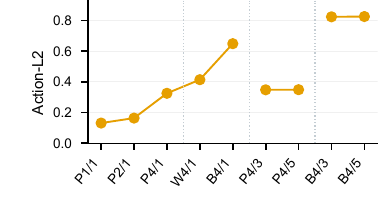}}%
        \makebox[0pt][r]{\raisebox{.5\ht0}{\footnotesize\textbf{(d)}\hspace{0.35em}}}\box0
    \end{minipage}
    \par
    \caption{\textbf{Action-response diagnostics.} \textbf{(a)} FastWAM Action-L2 versus SR across six patches ($r=-0.956$, $\rho=-0.714$). \textbf{(b)} One-shot and persistent exposure; dashed/solid denote local/trajectory responses. \textbf{(c)} LingBot-VA response over history length (log scale). \textbf{(d)} Cosmos $\sigma$-normalized cumulative Action-L2. P/W/B denote primary/wrist/both views; suffixes give patched-frame and joint-step counts. Protocols: Sections~\ref{app:action_surrogate}, \ref{app:e1_fixed}--\ref{app:cosmos_action}.}
    \label{fig:closed_loop_diagnostics}
    \label{fig:baseline_action_primary}
    
\end{figure}

\begin{samepage}
\section{Closed-Loop and Cross-Architecture Diagnostics}
\label{app:architecture}
\label{sec:closed_loop_diagnostics}

\subsection{Action discrepancy versus closed-loop success}
\label{app:action_surrogate}

Main-text Figure~\ref{fig:closed_loop_diagnostics}(a) remeasures the six primary baseline patches at a uniform runtime. Seed-42 patches share ten task-balanced observations with matched input, text, proprioception, and action noise; inference is BF16 with ten action steps and FP32 reduction. Two numerical repeats yield 120 measurements. We report repeat 0 and use the same patches' Eval500 success rates from Table~\ref{tab:main_attack_comparison}.

For normalized chunks $U_i^p,U_i^0\in\mathbb{R}^{H\times d}$, $H=32$, $d=7$, the diagnostic is
\begin{equation}
D_{\mathrm{chunk}}=\frac{1}{10}\sum_{i=1}^{10}
\frac{\sum_{h=1}^{H}\lVert U_{i,h}^p-U_{i,h}^0\rVert_2}{\sqrt{Hd}}.
\label{eq:baseline_chunk_action_l2}
\end{equation}
Equation~\ref{eq:baseline_chunk_action_l2} sums per-step action discrepancies over the full predicted chunk, including the unexecuted suffix. Equation~\ref{eq:e1_action_rms}, by contrast, uses RMS discrepancy for the cross-replan diagnostic.
\par\end{samepage}

For the six heterogeneous baselines, full-precision values give Pearson $r=-0.956$ and Spearman $\rho=-0.714$; three of 15 method pairs invert Action-L2 and Policy-SR order. Thus, Action-L2 is informative but does not fully rank attacks across different objectives. Deployment and fixed-input checks appear in Section~\ref{app:e1_fixed} and~\ref{app:cosmos_action}.

\paragraph{Layerwise association and interpretation}
Across 30 matched BS=1, 20k layers, Action-L2 is negatively associated with Eval200 SR ($r=-0.912$; $\rho=-0.880$; Table~\ref{tab:fastwam_action_l2_all30}). This complements the six-baseline diagnostic, where 3/15 pairwise orders reverse (Figure~\ref{fig:closed_loop_diagnostics}(a)); together, the results show that action discrepancy is informative but does not replace closed-loop evaluation.

\paragraph{Repeated exposure sustains disruption beyond the initial action}
For the FastWAM Top-3 patch, removing a one-shot patch largely restores same-state actions but not the resulting trajectory; persistent exposure increases trajectory RMS (Figure~\ref{fig:closed_loop_diagnostics}(b)). Changed actions and observations can therefore sustain divergence without an internal history cache (Section~\ref{app:e1_fixed}).

\paragraph{Broader internal loss targets do not consistently strengthen attacks}
Broader LingBot/Cosmos loss targets show no consistent singleton-attack gain (Table~\ref{tab:target_scope}). LingBot history changes K/V and action responses differently, while Cosmos Action-L2 varies with patched input extent and solver settings (Figure~\ref{fig:closed_loop_diagnostics}(c,d); Section~\ref{app:lingbot_history} and~\ref{app:cosmos_action}).

\subsection{FastWAM cross-replan deployment diagnostic}
\label{app:e1_fixed}

E1 uses the primary Top-3 patch (L29/L02/L03) on LIBERO-10. Trial 0 supplies matched initial states across clean, one-shot, and persistent conditions (environment/policy seed 42). One-shot patches R0; persistent deployment patches each current observation. The diagnostic has 30 trajectories and 1,325 replans.

\paragraph{Local versus closed-loop action discrepancy}
Let $U^c_{tr}\in\mathbb{R}^{32\times7}$ be the full normalized action chunk on condition $c$'s trajectory for task $t$, replan $r$, and let $\widetilde U^c_{tr}$ be the unpatched paired forward at that same current state. Let $U^0_{tr}$ be the actual action chunk from the separately executed clean branch at the same replan index. We report
\begin{equation}
d^{\mathrm{local}}_{tr,c}=\frac{\lVert U^c_{tr}-\widetilde U^c_{tr}\rVert_F}{\sqrt{32\cdot7}},\qquad
d^{\mathrm{loop}}_{tr,c}=\frac{\lVert U^c_{tr}-U^0_{tr}\rVert_F}{\sqrt{32\cdot7}}.
\label{eq:e1_action_rms}
\end{equation}
Figure~\ref{fig:closed_loop_diagnostics}(b) uses full-chunk normalized-action RMS, including the unexecuted suffix; the first $10\times7$ denormalized entries are executed.

Each R0--R4 point averages ten tasks; Figure~\ref{fig:closed_loop_diagnostics}(b) distinguishes one-shot local recovery from persistent disruption and trajectory divergence. After R0, the one-shot local discrepancy returns to the scale of the clean same-state discrepancy.

\subsection{LingBot fixed-input history and cache-aware attack}
\label{app:lingbot_history}

For fixed Tail-AV Top-3 inputs, increasing history length from R1 to R15 decreases mean relative K/V-L2 (0.3658/0.3519/0.3380/0.3276/0.3198 at R1/R2/R4/R8/R15), whereas normalized Action-L2 is non-monotonic (0.2649/1.0401/0.9335/0.5250/0.6796; Figure~\ref{fig:closed_loop_diagnostics}(c)). The action response peaks at R2, so the video-side cache displacement and action response do not move in lockstep. Relative K/V-L2 is normalized by the clean representation and does not isolate changes in the perturbation numerator or token span.

A separate cache-aware L05 patch backpropagates through R1--R4 history for 10k exposures. In this condition, clean and attacked SR are 94\% and 82\%; Task 4 accounts for 18 additional failures.

\subsection{Cosmos fixed-patch input sensitivity}
\label{app:cosmos_action}

Figure~\ref{fig:closed_loop_diagnostics}(d) uses the fixed 10k Top-5 patch (L14/L10/L01/L26/L19). Nine conditions and one input per task give 90 shared-noise pairs.

For normalized action chunks $a^p,a^0\in\mathbb{R}^{16\times7}$, the metric is
\begin{equation}
D_{\mathrm{act}}=\frac{1}{\sqrt{16\cdot7}}\sum_{t=1}^{16}
\left\|(a^p_t-a^0_t)/\sigma_{\mathrm{action}}\right\|_2,
\qquad
\sigma_{\mathrm{action}}=\frac{2\,\mathrm{std}(a)}{a_{\max}-a_{\min}}.
\label{eq:cosmos_cumulative_action_l2}
\end{equation}
All seven dimensions use dataset statistics and equal task averaging. This sigma-normalized cumulative reduction differs from FastWAM RMS.

With one joint step, patching one versus four primary-view frames raises $D_{\mathrm{act}}$ from 0.131 to 0.326; at four frames, wrist-only and both-view patching give 0.414 and 0.649 (Figure~\ref{fig:closed_loop_diagnostics}(d)). Increasing joint steps from one to three or five changes the primary-only four-frame response only to 0.348/0.349, whereas the both-view response reaches 0.825/0.827. Within these fixed inputs, patched view/frame coverage matters more than extra joint steps for the primary-only condition. These measurements quantify fixed-input action sensitivity.

Here $R$ counts patched current-slot frames (not history length), $S$ counts Hybrid-EDM joint steps, and diagnostic noise uses base seed 42 with per-input seed $4200000+\mathrm{sample\_index}$.

\FloatBarrier

\section{PASA-KV Guard}
\label{app:guard}
\label{sec:guard_results}

PatchGuard and PatchCleanser address adversarial patch robustness in image classification~\citep{xiang2021patchguard,xiang2022patchcleanser}. PASA-KV Guard instead intervenes on the visual K/V contributions read by action queries.

\subsection{Layer selection and clean calibration}

Thirty singleton Probe1000 patches are scored on ten paired task inputs with shared clean/attack noise. For normalized chunks $a^p_{\ell t},a^0_t\in\mathbb{R}^{H\times D}$,
\begin{equation}
s^{\mathrm{Guard}}_\ell=\frac{1}{10}\sum_{t=0}^{9}
\frac{\left\|(a^p_{\ell t}-a^0_t)/\sigma_{\mathrm{action}}\right\|_2}{\sqrt{HD}}.
\label{eq:guard_selection_score}
\end{equation}
Here $H=32$, $D=7$, and $\sigma_{\mathrm{action}}$ is the dataset's per-dimension global standard deviation mapped through the min/max normalizer. Descending scores select L02/L01/L06/L00/L12 independently of Tail-AV; All30 and Top5 thus test broad and targeted mitigation. One clean trajectory per task sets thresholds: the 99.5th percentile is computed within task and then averaged, giving $C_{\ell h}$ for visual-attention dominance and $T_{\ell h}$ for local contribution.

\begin{table}[h!]
    \centering
    \caption{Frozen high-response candidates selected from all 30 FastWAM layers by normalized paired Action-L2. Each layer has ten task-balanced paired records from an equal-budget Probe1000.}
    \label{tab:fastwam_guard_selected_layers}
    \footnotesize
    \begin{tabular}{@{}clcc@{}}
        \toprule
        Rank & Layer & Normalized paired Action-L2 & 95\% bootstrap CI \\
        \midrule
        1 & L02 & 0.47284 & $[0.36650, 0.58085]$ \\
        2 & L01 & 0.25876 & $[0.19552, 0.33376]$ \\
        3 & L06 & 0.12833 & $[0.09933, 0.15727]$ \\
        4 & L00 & 0.09233 & $[0.07501, 0.11057]$ \\
        5 & L12 & 0.08447 & $[0.06727, 0.10072]$ \\
        \bottomrule
    \end{tabular}
\end{table}

\subsection{Guard equations}
\label{app:guard_equations}

At each active layer/head, Guard applies the same token-wise rule; Targeted uses five selected layers and All30 uses all layers. With $D_j=\max_i A_{ij}$,
\begin{equation}
g^K_j=\sigma\!\left((D_j-C_{\ell h})/\tau\right),\qquad
K'_j=K_j(1-\lambda_K g^K_j).
\end{equation}
It recomputes attention with $K'_j$, compresses the visual tail, and normalizes each query row:
\begin{equation}
\widehat A_{ij}=C_{\ell h}\tanh(A'_{ij}/C_{\ell h}),\qquad
\overline A_{ij}=\widehat A_{ij}/\sum_k\widehat A_{ik}.
\end{equation}
The bound $\widehat A_{ij}\leq C_{\ell h}$ holds before, not after, row normalization; Guard compresses tails rather than imposes a hard final cap.
Finally, with $I'_{ij}=\overline A_{ij}\lVert V_j\rVert_2$,
\begin{equation}
g^V_{ij}=\sigma\!\left((I'_{ij}-T_{\ell h})/\tau\right),\qquad
y_i^{\mathrm{guard}}=\sum_j\overline A_{ij}(1-\lambda_Vg^V_{ij})V_j.
\end{equation}
We use $\tau=0.02$ and $\lambda_K=\lambda_V=0.50$ in \texttt{kv\_joint} mode. The differentiable rule uses no patch coordinates, position prior, or attack label.

\subsection{Fixed Tail-AV patch results}
\label{app:guard_fixed_results}

We evaluate Guard against fixed, non-adaptive patches. Table~\ref{tab:fastwam_guard_main} reports pooled results and Table~\ref{tab:fastwam_guard_fixed_patches} the three 200-episode fixed patches: L29/L02/L03, then L01, then L24. Targeted Guard protects all 24 heads at L02/L01/L06/L00/L12 under fixed calibration.

\begin{table}[H]
\centering
\caption{\textbf{Pooled Guard results on fixed Tail-AV attacks.} SR and recovery are percentages. Attacked SR pools three patches (600 episodes); clean SR uses 200 episodes per deployment. Recovery is the fraction of the attack-induced SR loss restored, relative to the 93.5\% undefended clean reference. Per-patch results: Table~\ref{tab:fastwam_guard_fixed_patches}.}
\label{tab:fastwam_guard_main}
\footnotesize
\setlength{\tabcolsep}{7pt}
\begin{tabular}{@{}lrrr@{}}
\toprule
Deployment & Clean SR & Attacked SR & Recovery \\
\midrule
No defense & 93.50 & 1.33 & -- \\
Guard (30 layers) & 53.00 & 49.67 & 52.44 \\
Guard targeted (5) & 93.50 & 87.50 & 93.49 \\
\bottomrule
\end{tabular}
\end{table}

\begin{table}[H]
\centering
\caption{\textbf{Per-patch Guard results.} Policy SR (\%) and success counts, with 200 episodes per patch/deployment. No-defense results use the fixed patch family evaluated for Guard.}
\label{tab:fastwam_guard_fixed_patches}
\footnotesize
\setlength{\tabcolsep}{5pt}
\begin{tabular}{@{}lrrrrrr@{}}
\toprule
 & \multicolumn{2}{c}{No defense} & \multicolumn{2}{c}{Guard (30)} & \multicolumn{2}{c}{Guard targeted (5)} \\
\cmidrule(lr){2-3}\cmidrule(lr){4-5}\cmidrule(lr){6-7}
Input & SR & Count & SR & Count & SR & Count \\
\midrule
Clean & 93.50 & 187/200 & 53.00 & 106/200 & 93.50 & 187/200 \\
\midrule
Tail-AV Top-3 & 0.00 & 0/200 & 52.50 & 105/200 & 85.50 & 171/200 \\
Tail-AV Top-4 & 0.00 & 0/200 & 40.50 & 81/200 & 89.00 & 178/200 \\
Tail-AV Top-5 & 4.00 & 8/200 & 56.00 & 112/200 & 88.00 & 176/200 \\
\midrule
Pooled attacks & 1.33 & 8/600 & 49.67 & 298/600 & 87.50 & 525/600 \\
\bottomrule
\end{tabular}
\end{table}

The recovery fraction $(\mathrm{SR}_{\mathrm{guard}}-\mathrm{SR}_{\mathrm{attack}})/(93.5-\mathrm{SR}_{\mathrm{attack}})$ is 52.44\% for All30 and 93.49\% for Targeted5. Compared with All30, targeted Guard improves both clean and attacked SR while matching the undefended clean reference.

Separate Eval200 clean controls give All30 53.0\% (106/200; Wilson 95\% CI 46.09--59.80\%) and Targeted5 93.5\% (187/200; 89.20--96.16\%). Clean controls are counted once across the patch family; calibration trajectories are excluded.

\FloatBarrier
\begingroup
\setlength{\intextsep}{6pt plus 2pt minus 2pt}
\setlength{\textfloatsep}{10pt plus 2pt minus 2pt}
\section{Real-robot deployment and results}
\label{app:real_robot}
\label{sec:real_robot}

\paragraph{Platform and patch protocol}
We tested update-10,500 FastWAM on a 6-DoF AgileX Piper with head and wrist RGB cameras. Its four-task suite covers pen-in-cup, carrot-in-bowl, second-drawer placement and closing, and bowl stacking. Figure~\ref{fig:piper_task_suite} pairs each initial scene with an annotated representative frame from the corresponding ground-truth rollout. The $224\times224$ views were concatenated side by side; at each replan, the digital patch was overlaid on the current head view while the wrist view remained unchanged.

\begin{figure}[!htbp]
\centering
\includegraphics[width=\textwidth]{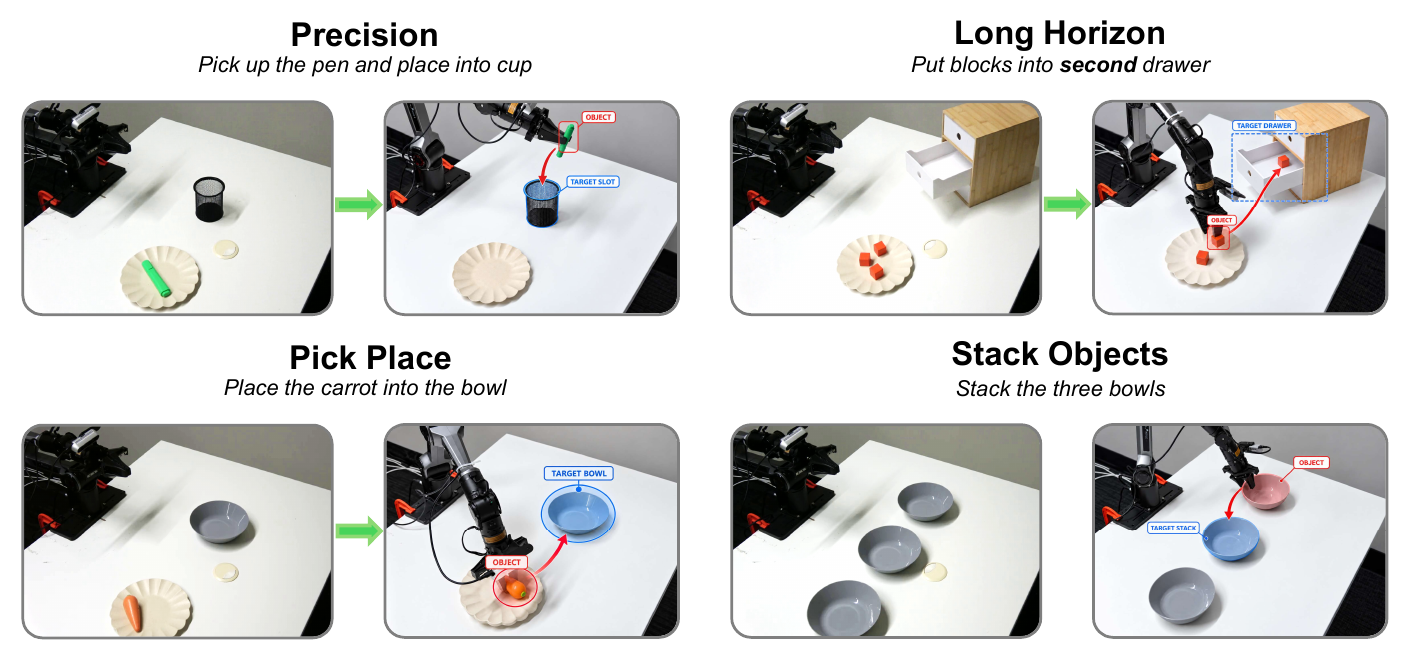}
\caption{\textbf{Piper FastWAM task scenes and reference rollouts.} Each pair shows the initial scene (left) and a representative annotated frame from the provided ground-truth rollout (right). The manipulated object is marked in red and its target slot or placement site in blue. Tasks are pen-in-cup, second-drawer placement, carrot-in-bowl, and bowl stacking; closed-loop attack results appear in Figure~\ref{fig:piper_tasks}.}
\label{fig:piper_task_suite}
\end{figure}

\paragraph{PASA-KV construction}
Probe100 scored all 30 video layers using task-balanced samples and one held-out input per task. Top-3 selected L04/L27/L00; Top-5 added L02/L01. For each Top-$k$, one fresh patch was trained with four-task-balanced samples and shared across tasks at deployment. The $50\times50$ RGB patches used 10,000 BS1 AdamW ascent updates (learning rate $5\times10^{-4}$, seed 42), without transformation sampling. Each patch was placed at $(5,170)$ in the head view; only RGB pixels were optimized. Top-4 was trained but not physically evaluated.
\begin{figure}[!htbp]
\centering
\includegraphics[width=0.82\textwidth,trim=0bp 24bp 0bp 8bp,clip]{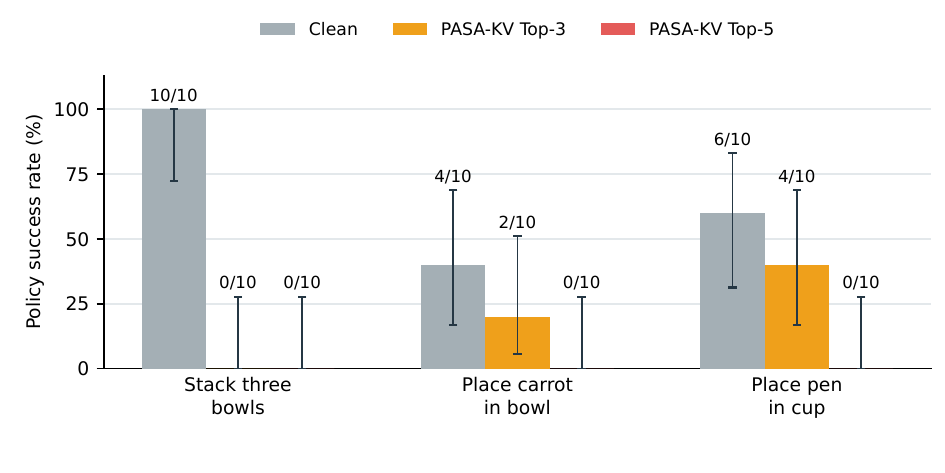}
\caption{\textbf{Piper real-robot attack effects across three tasks.} Bars show observed closed-loop task success; labels give successes/10. Error bars show two-sided 95\% Wilson binomial confidence intervals over ten trials per condition. Top-3 and Top-5 use separately trained task-shared patches. The drawer task, whose clean SR is 0/10, is excluded from attack comparison. The input perturbation is a digital head-camera overlay.}
\label{fig:piper_tasks}
\end{figure}

\paragraph{Evaluation coverage}
Each physical condition used 10 in-distribution trials. Figure~\ref{fig:piper_tasks} reports clean, Top-3, and Top-5 on three tasks; clean FastWAM scored 0/10 on the second-drawer task, with no attack comparison. Figure~\ref{fig:piper_baselines} gives the shared stacking-task results. Baselines used 10,000 updates without EOT: DP-Attacker used $45\times45$ at $(5,175)$, DRIFT $32\times32$ at $(5,188)$, and the other four $50\times50$ at $(5,170)$.

\begin{figure}[!htbp]
\centering
\includegraphics[width=0.83\textwidth]{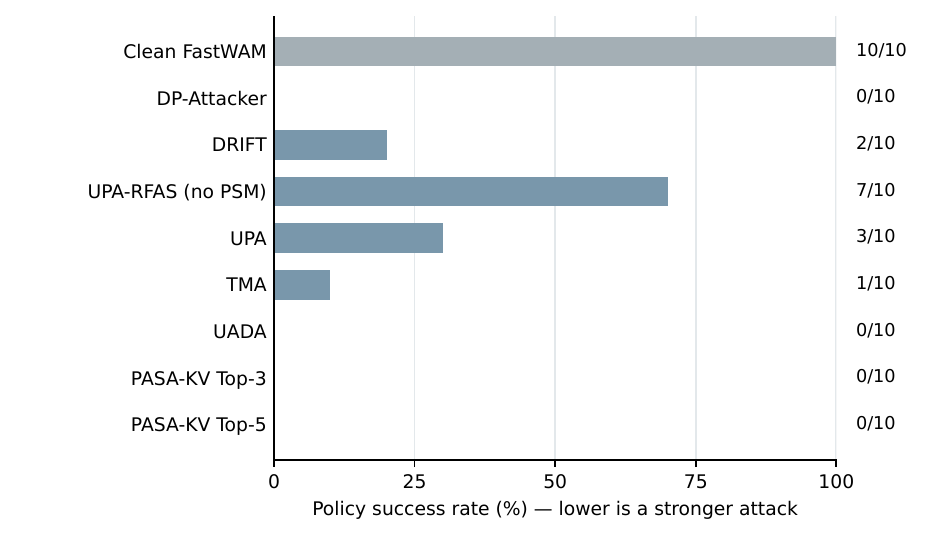}
\caption{\textbf{Bowl-stacking success under PASA-KV and baseline attacks.} Each method has ten in-distribution physical trials. Bars show observed task success; labels give successes/10. Each method uses its own patch geometry, as specified in Section~\ref{sec:real_robot}. Lower SR means a stronger attack.}
\label{fig:piper_baselines}
\end{figure}

On bowl stacking, Top-3 reaches 0/10, matched only by DP-Attacker and UADA among the six baselines. Top-5 yields 0/10 on all three evaluated tasks with nonzero clean success.
\FloatBarrier
\endgroup

\section{Discussion}
\label{sec:discussion}

The cross-system comparison evaluates one checkpoint per architecture using the WAM adaptations described in Appendix~\ref{app:baselines}. Real-robot evaluation covers three tasks; the full baseline comparison is limited to bowl stacking.

\section{Conclusion}
\label{sec:conclusion}

We introduce PASA-KV, which uses short action-facing probes to select visual K/V routes before training a fresh attack. Matched controls validate readout-informed selection. Across architectures, Top-5 reaches 0.0\%, 87.2\%, and 89.2\% SR on FastWAM, LingBot-VA, and Cosmos Policy; recorded construction costs are lower on FastWAM and LingBot-VA. On a separate Piper checkpoint, one task-shared Top-5 patch yields 0/10 success on three clean-capable tasks. Together, the results distinguish representation perturbability, action-facing response, and closed-loop vulnerability.

\Needspace{9\baselineskip}
\section*{Declarations}
\addcontentsline{toc}{section}{Declarations}
\subsection*{AI use statement}

Generative AI tools were used to assist with literature search and synthesis, language editing and writing refinement, and the creation or refinement of scientific figures. AI-generated suggestions and materials were reviewed, revised, and verified by the authors before inclusion in the manuscript. The authors take responsibility for the final content of this work, including text and figures produced with the aid of generative AI.

\subsection*{Ethics statement}

This work studies adversarial attacks to characterize and reduce security risks in robot-control models. The evaluation includes benchmark simulation and controlled tabletop rollouts on a 6-DoF Piper robot. Physical trials were limited to the predefined task suite. No human-subject study was conducted. The work is intended to support robustness research. Because attack methods can be dual-use, we report threat-model constraints, defense failures, and the scope of the non-adaptive evaluation. Any public release of attack artifacts should be reviewed for responsible disclosure and misuse risk.

\subsection*{Reproducibility statement}
Sections~\ref{app:registered_threats} and~\ref{app:reduced_step_controls} specify the primary threat models and clean solver controls. Sections~\ref{app:optimization} and~\ref{app:selectors} give construction budgets, gradient paths, scoring definitions, and capture interfaces. The counterfactual mechanism protocol, Probe100 predictive analysis, and matched selector controls appear in Sections~\ref{app:kv_counterfactual}, \ref{app:probe_prediction}, and~\ref{app:selector_controls}. Sections~\ref{app:architecture}, \ref{app:guard}, and~\ref{app:real_robot} give action-response diagnostics, Guard equations and calibration, and Piper deployment coverage. Appendix~\ref{app:reproducibility} retains exact checkpoint identity, patch lineage, evaluation denominators, and complete rankings; Appendices~\ref{app:kv_mechanism}, \ref{app:layerwise}, and~\ref{app:baselines} retain the derivation, full mechanism and layerwise ledgers, and baseline adaptations.

The primary attack protocol is indexed by label $42$, while singleton training configurations record seed $20260801$. These identifiers refer to different experiment groups. Section~\ref{app:selector_controls} specifies the fixed-initialization protocol for the repeated runs, and each result is linked to its own victim, patch, and evaluation records. The trained victim checkpoint (\texttt{step\_020000.pt}), code, and machine-readable result ledgers are planned for release. Release details will be announced separately.

\FloatBarrier
\bibliographystyle{plainnat}
\bibliography{references}
\appendix
\section{Reproducibility Details and Complete Rankings}
\label{app:reproducibility}
This appendix lists checkpoint identities, patch provenance, evaluation sets, and complete layer rankings.
\begin{table}[!htbp]
\centering
\caption{\textbf{Production Tail-AV selector scores and ranking on the evaluated FastWAM checkpoint.} Descending Eq.~\ref{eq:tail_av_score}, with layer-ID tie-breaking; scores rounded to six decimals. Top-3/4/5 prefixes give the selected attack sets.}
\label{tab:selector_full_rankings}
\footnotesize
\setlength{\tabcolsep}{4pt}
\begin{tabular}{@{}rlr@{\hspace{1.2em}}rlr@{\hspace{1.2em}}rlr@{}}
\toprule
Rank & Layer & Score & Rank & Layer & Score & Rank & Layer & Score \\
\midrule
1 & L29 & $0.049694$ & 11 & L14 & $0.009106$ & 21 & L28 & $-0.008567$ \\
2 & L02 & $0.043262$ & 12 & L19 & $0.007954$ & 22 & L10 & $-0.017586$ \\
3 & L03 & $0.032456$ & 13 & L16 & $0.004701$ & 23 & L26 & $-0.029588$ \\
4 & L01 & $0.027921$ & 14 & L17 & $0.004437$ & 24 & L22 & $-0.034651$ \\
5 & L24 & $0.026162$ & 15 & L15 & $0.002879$ & 25 & L09 & $-0.038692$ \\
6 & L00 & $0.023169$ & 16 & L18 & $0.002561$ & 26 & L06 & $-0.043447$ \\
7 & L20 & $0.021821$ & 17 & L13 & $0.001807$ & 27 & L05 & $-0.047775$ \\
8 & L12 & $0.019057$ & 18 & L11 & $-0.000352$ & 28 & L07 & $-0.049264$ \\
9 & L27 & $0.014059$ & 19 & L21 & $-0.000550$ & 29 & L08 & $-0.061348$ \\
10 & L25 & $0.012028$ & 20 & L23 & $-0.008298$ & 30 & L04 & $-0.075180$ \\
\bottomrule
\end{tabular}
\end{table}

\begin{table}[!htbp]
\centering
\caption{\textbf{Experiment map.} Patch lineage is listed for each evidence block. Eval200/500 denote 20/50 online trials per task.}
\label{tab:experiment_map}
\footnotesize
\setlength{\tabcolsep}{4pt}
\begin{tabular}{@{}p{0.23\textwidth}p{0.36\textwidth}p{0.36\textwidth}@{}}
\toprule
Evidence & Patch relationship & Endpoint and role \\
\midrule
Primary comparison & Separately trained Top-3, Top-4, and Top-5 patches per model & Eval500; baseline comparison and target-count tradeoff \\
Matched selection & Primary FastWAM Top-3 plus separately trained controls & Common Eval200; isolates layer-set choice \\
Training repeats & Same layer sets and exact initial pixels; varied final-training seed and shuffled sample order & Common Eval200; fixed-set training sensitivity \\
Layerwise analysis & Separate singleton-target patches & SR trajectories and paired internal responses; target characterization \\
Input diagnostics & Patch specified per panel & Paired action/representation responses; FastWAM also has 30 diagnostic rollouts \\
Guard & Separate fixed Top-3/4/5 family & 200 episodes per patch and deployment; mitigation \\
Real robot & Piper-specific fresh Top-3/5 patches & Three shared tasks, 10 trials per task; physical closed-loop control \\
\bottomrule
\end{tabular}
\end{table}

\begin{table}[!htbp]
\centering
\caption{\textbf{FastWAM victim training configuration.} Values describe the frozen victim model used for the FastWAM experiments.}
\label{tab:fastwam_victim_config}
\footnotesize
\setlength{\tabcolsep}{3pt}
\begin{tabular}{@{}p{0.19\textwidth}p{0.76\textwidth}@{}}
\toprule
Item & Documented record \\
\midrule
Hardware and distributed training & 8 MetaX GPUs; DeepSpeed ZeRO-2; FP32; seed 42; per-GPU batch 8; gradient accumulation 2; global batch 128 \\
Data & LIBERO-Spatial, LIBERO-Object, LIBERO-Goal, and LIBERO-10; 40 tasks; 1,712 demonstrations; 277,713 training frames \\
Schedule & 10 planned epochs; 21,700 planned optimizer updates; save/eval/log frequencies 2,000/200/10 updates \\
Optimizer and schedule & AdamW; learning rate $10^{-4}$; betas $(0.9,0.95)$; epsilon $10^{-8}$; weight decay 0.01; gradient clipping 1.0; 1,085-update 5\% warm-up; cosine decay to $10^{-6}$; no EMA or early stopping \\
Model & \texttt{Wan-AI/Wan2.2-TI2V-5B}; approximately 5.00B video-expert and 1.02B action-expert parameters; 30-layer MoT; video/action/proprioception trainable, VAE and other modules frozen; both loss weights 1.0 \\
ModelScope revision & \texttt{c2272aadfec2d95cea869a4e748d5e28275621c6} \\
Input and inference & Two $512\times512$ views resized to $224\times224$ and concatenated to $224\times448$; 9 video frames; 32-step 7-D action chunk; 8-D proprioception; min/max normalization; no random image augmentation; attack inference BF16 with 10 action-denoising steps \\
Validation scope & $\texttt{val\_set\_proportion}=0$; periodic evaluation reused the training set, while paper Policy SR uses separate simulator rollouts. \\
\bottomrule
\end{tabular}
\end{table}

\subsection{Matched selector-control layer sets}
\definecolor{groupEarly}{HTML}{D45A54}
\definecolor{groupIntermediate}{HTML}{DAA036}
\definecolor{groupOther}{HTML}{8B969E}
\newcommand{\earlydot}{\raisebox{0.1ex}{\textcolor{groupEarly}{$\bullet$}}}
\newcommand{\intermediatedot}{\raisebox{0.1ex}{\textcolor{groupIntermediate}{$\bullet$}}}
\newcommand{\otherdot}{\raisebox{0.1ex}{\textcolor{groupOther}{$\bullet$}}}

\begin{table}[!htbp]
\centering
\caption{\textbf{Matched FastWAM selection controls.} Same joint K/V loss, three targets, geometry, and 10k exposures; Eval200 (10 tasks $\times$ 20 trials). Dots denote retrospective response classes.}
\label{tab:selector_efficacy_controls_appendix}
\footnotesize
\setlength{\tabcolsep}{5pt}
\begin{tabular}{@{}lllrr@{}}
\toprule
Selection rule & Layer set & Class & Successes & Policy SR (\%) \\
\midrule
Probe-$J^{KV}$ & L27/L23/L28 & \otherdot/\otherdot/\otherdot & 95/200 & 47.5 \\
Evenly spaced & L00/L15/L29 & \earlydot/\intermediatedot/\earlydot & 37/200 & 18.5 \\
Random-1 & L27/L18/L16 & \otherdot/\intermediatedot/\intermediatedot & 173/200 & 86.5 \\
Random-2 & L12/L23/L00 & \otherdot/\otherdot/\earlydot & 127/200 & 63.5 \\
Random-3 & L20/L22/L00 & \otherdot/\otherdot/\earlydot & 50/200 & 25.0 \\
\midrule
Tail-AV & L29/L02/L03 & \earlydot/\earlydot/\earlydot & 3/200 & \textbf{1.5} \\
\bottomrule
\end{tabular}
\end{table}

\FloatBarrier
\section{Additional K/V Derivation and Complete Mechanism Evidence}
\label{app:kv_mechanism}

\paragraph{First-order decomposition}
For an action query $q_i$ and a visual token $j$, let
\begin{equation}
    a_{ij}=\operatorname{softmax}_{j}\!\left(q_i^\top k_j/\sqrt d\right),
    \qquad y_i=\sum_j a_{ij}v_j.
\end{equation}
Holding $q_i$ fixed and perturbing only visual K/V gives the first-order expansion
\begin{equation}
\delta y_i\approx
\underbrace{\frac{1}{\sqrt d}\sum_{j\in\mathcal V}a_{ij}
    (q_i^\top\delta k_j)(v_j-y_i)}_{u_K\text{ (K-mediated routing)}}
+\underbrace{\sum_{j\in\mathcal V}a_{ij}\delta v_j}_{u_V\text{ (V-mediated content)}}.
\label{eq:appendix_first_order_kv}
\end{equation}
The K term depends on query alignment and value contrast, while the V term is gated by action-facing attention; raw representation displacement alone is therefore insufficient. The joint contribution contains a cross-term:
\begin{equation}
    \lVert u_K+u_V\rVert_2^2
    =\lVert u_K\rVert_2^2+\lVert u_V\rVert_2^2+2\langle u_K,u_V\rangle.
\label{eq:appendix_kv_cross_term}
\end{equation}
For reference, Cauchy--Schwarz gives the bounds
\begin{equation}
\begin{split}
\lVert u_K\rVert_2 &\leq \frac{\lVert q_i\rVert_2}{\sqrt d}
\left(\sum_j a_{ij}\lVert\delta k_j\rVert_2^2\right)^{1/2}
\left(\sum_j a_{ij}\lVert v_j-y_i\rVert_2^2\right)^{1/2},\\
\lVert u_V\rVert_2 &\leq \left(\sum_j a_{ij}\lVert\delta v_j\rVert_2^2\right)^{1/2}.
\end{split}
\end{equation}
Thus, global $D_K$ and $D_V$ do not alone characterize action-facing vulnerability. The cross-term in Eq.~(\ref{eq:appendix_kv_cross_term}) shows that K- and V-mediated readout changes can reinforce or oppose each other.

\subsection{Matched component-attack evidence}
The matched 10k-exposure sweep compares K-only, V-only, and joint K/V objectives under the same FastWAM protocol. Table~\ref{tab:fastwam_kv_complementarity} summarizes paired SR differences and strong-attack coverage ($\mathrm{SR}\leq50\%$) across the 30 layers. Complete component and architecture-wide sweeps are provided in Appendix~\ref{app:component_sweeps}.
\begin{table}[H]
    \centering
    \caption{\textbf{Paired objective comparisons across the 30 layers of the evaluated FastWAM checkpoint.} Lower Policy SR is a stronger attack. ``Strong attack'' denotes the descriptive coverage threshold SR $\leq 50\%$. Percentile intervals resample the 30 evaluated layers 20,000 times with seed 20260905.}
    \label{tab:fastwam_kv_complementarity}
    \footnotesize
    \setlength{\tabcolsep}{3pt}
    \begin{tabular}{@{}p{0.54\textwidth}r p{0.25\textwidth}@{}}
        \toprule
        Paired layer statistic & Estimate & 95\% interval \\
        \midrule
        K+V strictly stronger than K-only & 17/30 (56.7\%) & [40.0, 73.3]\% \\
        K+V strictly stronger than V-only & 17/30 (56.7\%) & [40.0, 73.3]\% \\
        Median $\mathrm{SR}_{K+V}-\mathrm{SR}_{K}$ & $-1.25$ pp & [$-3.75$, 0.50] pp \\
        Median $\mathrm{SR}_{K+V}-\mathrm{SR}_{V}$ & $-1.00$ pp & [$-3.50$, 0.25] pp \\
        Strong-attack layer coverage: K-only & 7/30 (23.3\%) & [10.0, 40.0]\% \\
        Strong-attack layer coverage: V-only & 11/30 (36.7\%) & [20.0, 53.3]\% \\
        Strong-attack layer coverage: K+V & 13/30 (43.3\%) & [26.7, 60.0]\% \\
        K+V retains strong attack in single-branch-asymmetric layers & 8/10 (80.0\%) & [50.0, 100.0]\% \\
        \bottomrule
    \end{tabular}
\end{table}

\begin{table}[H]
    \caption{\textbf{All-layer component-objective attacks on the evaluated FastWAM checkpoint.} Values are pooled closed-loop Policy SR (\%; lower is stronger) over independently trained single-layer patches. K-only and V-only denote end-to-end patch optimization targets. Bold marks the lowest SR.}
    \label{tab:component_attack_coverage}
    \centering
    \footnotesize
    \setlength{\tabcolsep}{3.8pt}
    \begin{tabular}{@{}lrrrr@{}}
        \toprule
        Model & Attn.-map & K only & V only & K+V \\
        \midrule
        FastWAM & 90.68 & 65.77 & 66.43 & \textbf{58.87} \\
        \bottomrule
    \end{tabular}
    \vspace{2pt}

    \begin{minipage}{0.98\columnwidth}
    \footnotesize
    \textit{Protocol.} All objectives use BS1 and 10k exposures per patch; each value pools 30 layers $\times$ 200 episodes (6,000). The complete three-model summary and layerwise values appear in Appendix~\ref{app:component_sweeps} (Table~\ref{tab:component_attack_coverage_full}).
    \end{minipage}
\end{table}

\subsection{Counterfactual readout and representative mechanism cases}
\label{app:kv_counterfactual}

The mechanism sweep covers 30 FastWAM layers, three optimization objectives, and 50 observations from ten task-balanced inputs. K-only and V-only denote optimization objectives, so their patches may alter both cache components. For every joint K$+$V patch, we fix $Q^0_{\rm action}$ and construct:
\begin{equation}
\begin{aligned}
Y_{00}&=Y(Q^0,K^0,V^0), & Y_{K0}&=Y(Q^0,K^p,V^0),\\
Y_{0V}&=Y(Q^0,K^0,V^p), & Y_{KV}&=Y(Q^0,K^p,V^p).
\end{aligned}
\end{equation}
Let $\Delta_K=Y_{K0}-Y_{00}$, $\Delta_V=Y_{0V}-Y_{00}$, $\Delta_{KV}=Y_{KV}-Y_{00}$, and $\Delta_{\rm int}=Y_{KV}-Y_{K0}-Y_{0V}+Y_{00}$. The identity $\Delta_{KV}=\Delta_K+\Delta_V+\Delta_{\rm int}$ has maximum relative residual $1.273\times10^{-7}$. Attention uses the full key set; visual columns $[0,98)$ are selected after softmax without renormalization. Clean/attacked forwards share noise, and values average within task then over tasks.

\begin{figure}[H]
\centering
\includegraphics[width=0.60\textwidth]{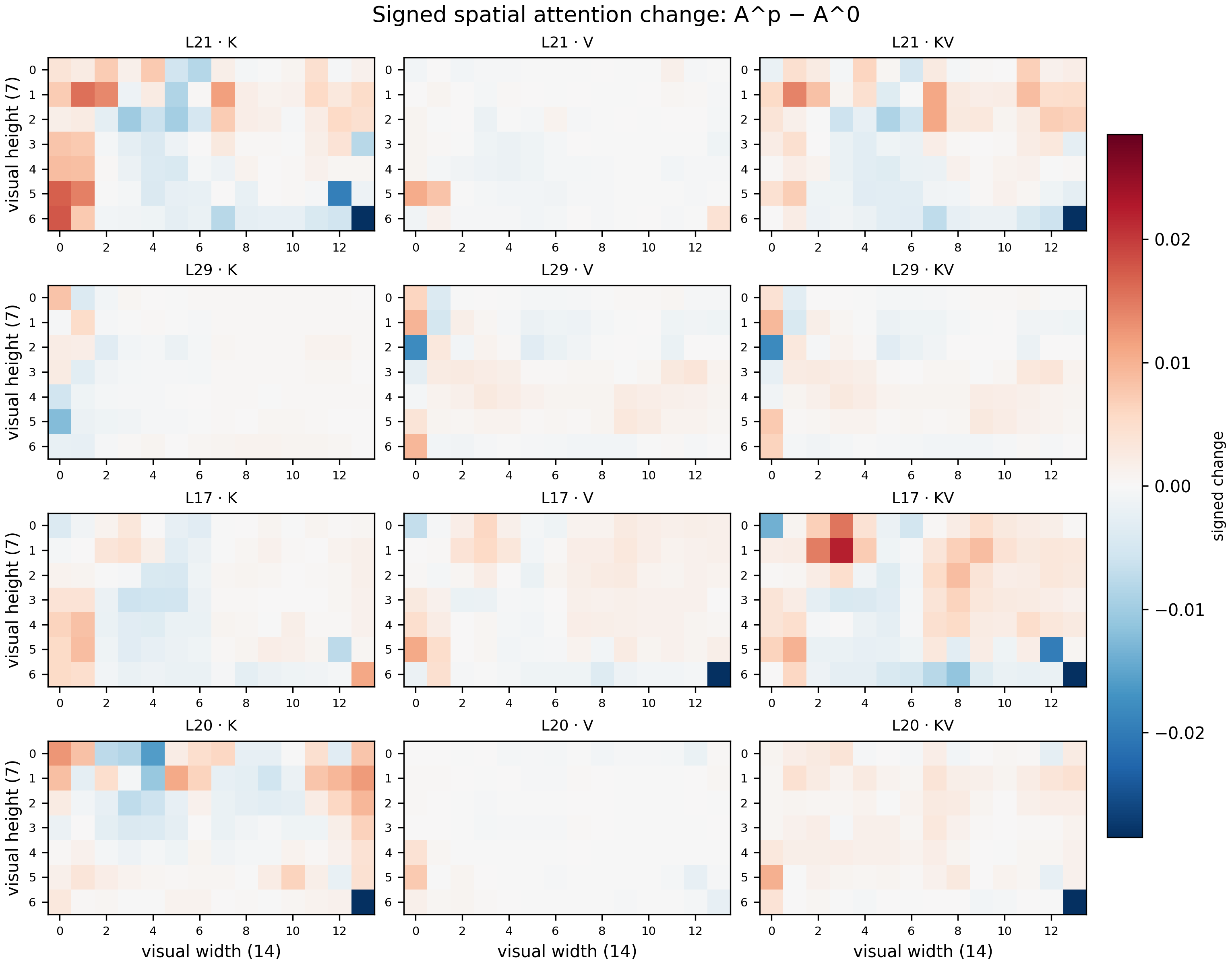}
\caption{\textbf{Spatial changes in action-facing attention on the evaluated FastWAM checkpoint.} Each panel shows the task-macro signed mean of $\Delta A_{ij}=A^p_{ij}-A^0_{ij}$, averaged over action queries and heads and mapped to the $7\times14$ visual-token grid after full-key softmax. Rows show L21/L29/L17/L20; columns show patches trained with K-only, V-only, and joint K$+$V objectives. K-only and V-only denote training objectives, not isolated cache-tensor interventions. Red/blue indicate increases/decreases.}
\label{fig:delta_A_spatial}
\end{figure}

Table~\ref{tab:fastwam_mechanism_full} gives the complete layerwise summary. For joint K$+$V, correlations with $\mathrm{AttackStrength}$ $=1-\mathrm{SR}/100$ are $0.316$ ($D_K$), $0.528$ ($D_V$), $0.530$ ($D_A$), and $0.332$ ($\lVert\Delta_{KV}\rVert$). Figures~\ref{fig:delta_A_spatial} and~\ref{fig:delta_influence_spatial} show where attention probabilities and attention-weighted visual-value contributions change across the four representative layers and three optimization objectives.

\paragraph{Representative mechanism cases}
Four cases illustrate layer dependence. L21 is K-dominant (K/V SR 6.0/91.0\%; mean $|\Delta A|$ 0.00511/0.00074); L29 is V-dominant (K/V/K$+$V SR 92.5/14.5/9.5\%; contribution change 0.01003/0.08218/0.08192); and L17 is joint-beneficial (K/V/K$+$V SR 66.0/91.5/11.5\%; mean $|\Delta A|$ 0.00207/0.00250/0.00534). L20 illustrates non-dominance: K-only reaches 0.0\% SR ($\lVert\Delta K\rVert=95.4$, $|\Delta A|=0.00449$), whereas K$+$V reaches 91.5\% ($\lVert\Delta_K\rVert=36.4$, $\lVert\Delta_{KV}\rVert=117.2$, $|\Delta A|=0.00251$). Thus, large internal displacement alone does not determine closed-loop failure.

\begin{figure}[!tbp]
    \centering
    \includegraphics[width=0.76\textwidth]{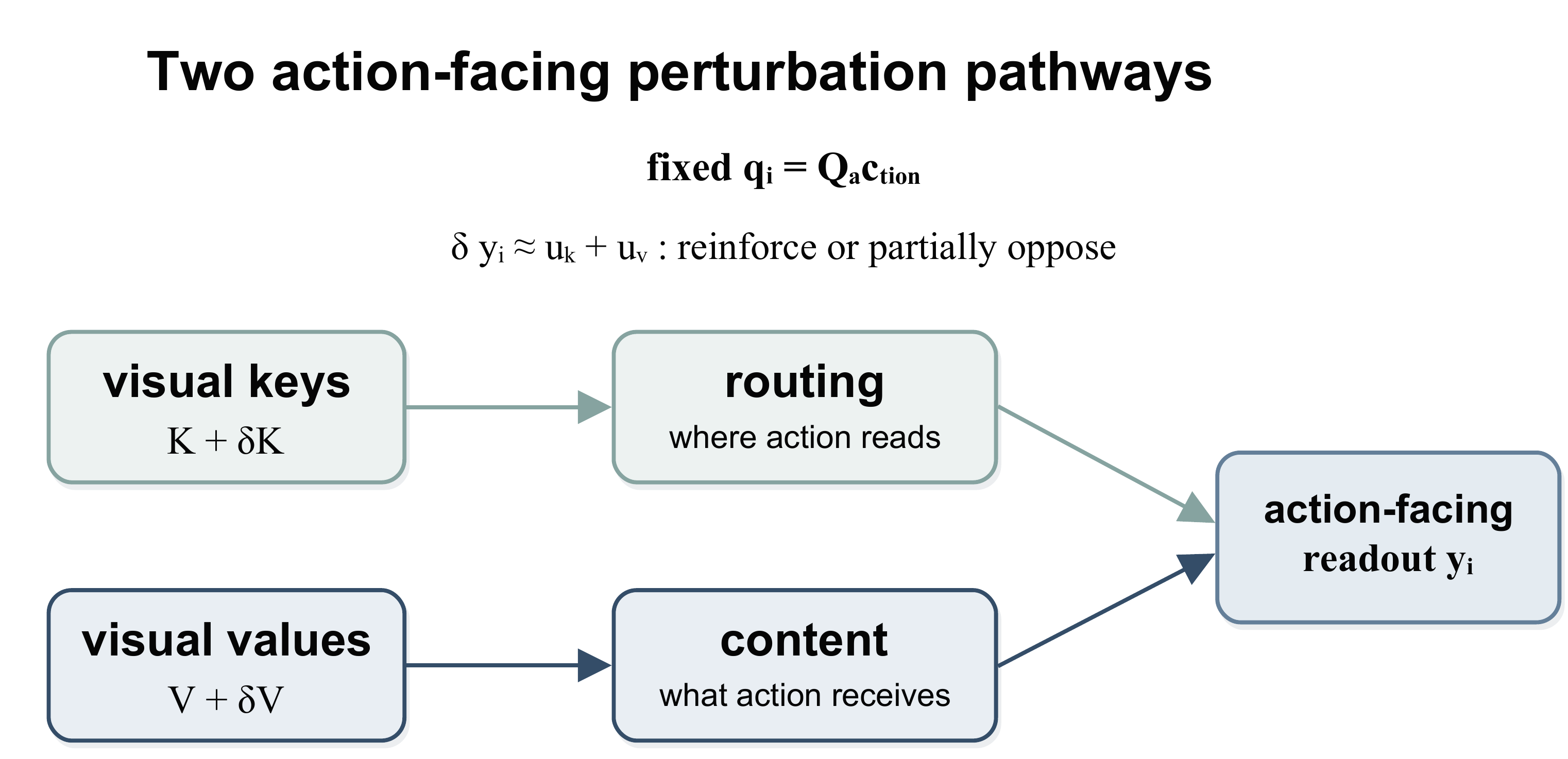}
    \caption{\textbf{Action-facing visual K/V pathways.} A fixed action query reads visual keys through attention and aggregates visual values into the action-facing readout. Key and value perturbations can alter routing and received content, respectively.}
    \label{fig:kv_pathway_schematic}
\end{figure}

\begin{figure}[!tbp]
    \centering
    \includegraphics[width=\textwidth]{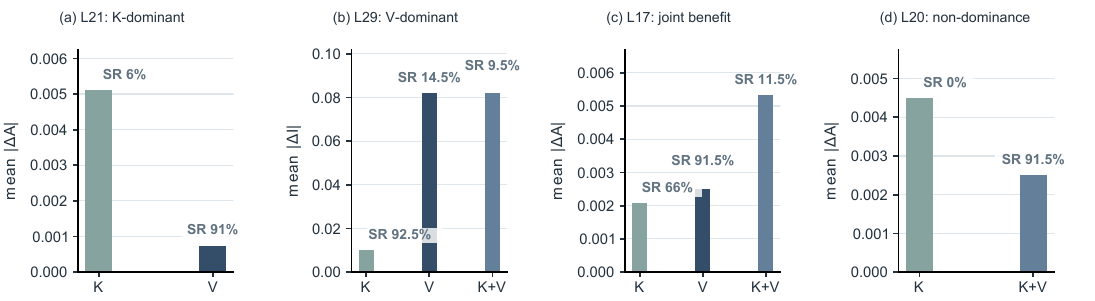}
    \caption{\textbf{Representative component-objective attack cases on the evaluated FastWAM checkpoint.} \textbf{(a)} L21, K-dominant. \textbf{(b)} L29, V-dominant. \textbf{(c)} L17, joint-beneficial. \textbf{(d)} L20, joint non-dominance. Bars show the objective contrasts discussed in the text; labels report singleton Policy SR.}
    \label{fig:kv_mechanism_cases}
\end{figure}

\begin{figure}[H]
\centering
\includegraphics[width=0.60\textwidth]{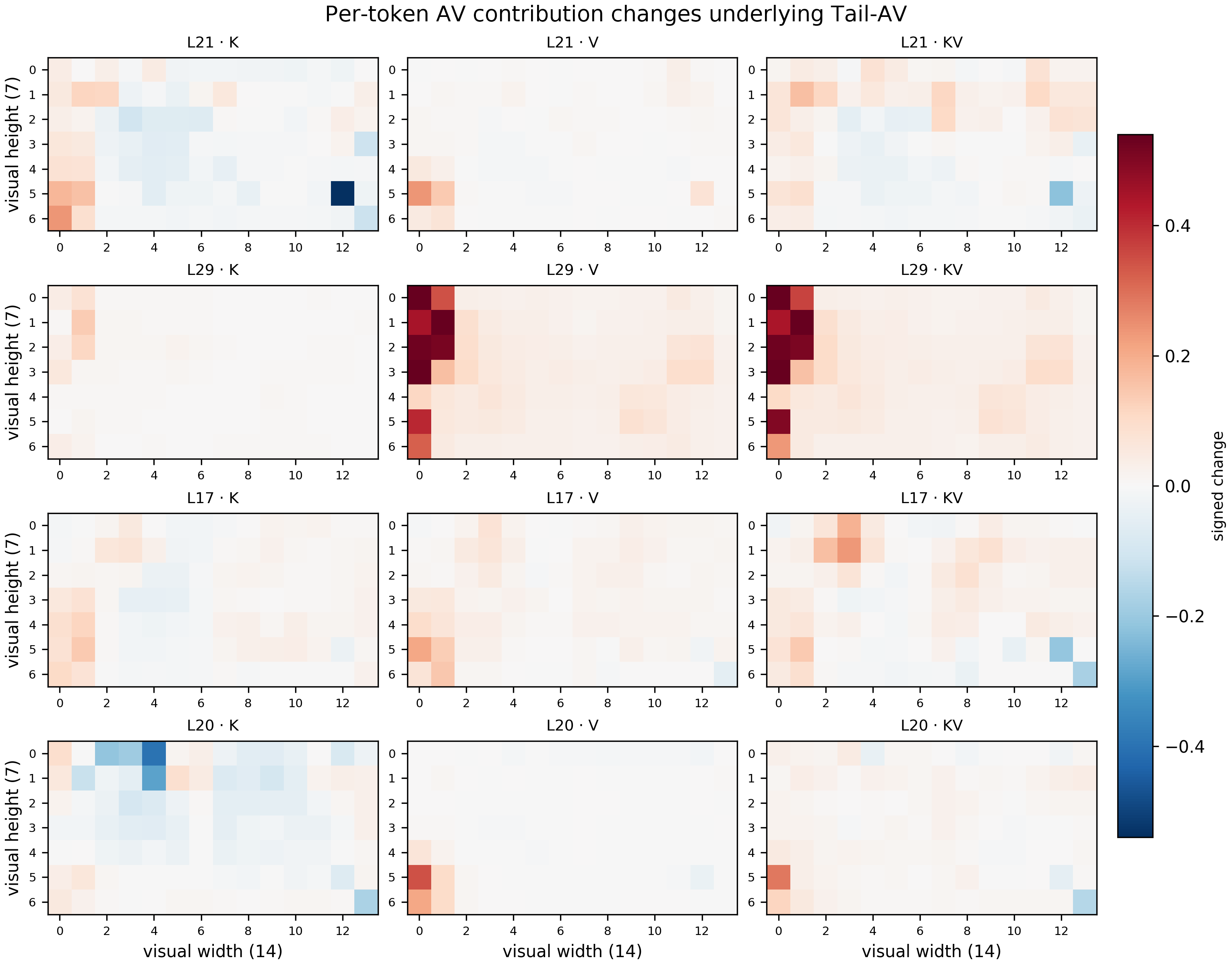}
\caption{\textbf{Spatial changes in action-facing visual contribution on the evaluated FastWAM checkpoint.} Each panel shows the signed task-macro mean over 50 observations of $\Delta I_{ij}=A^p_{ij}\lVert V^p_j\rVert_2-A^0_{ij}\lVert V^0_j\rVert_2$, averaged over action queries and heads on the $7\times14$ visual-token grid. Rows show L21/L29/L17/L20; columns show K-only, V-only, and joint K$+$V optimization objectives. K-only and V-only name training objectives, not isolated cache-tensor interventions. Red/blue indicate increases/decreases. These maps complement the quantitative case summary in Figure~\ref{fig:kv_mechanism_cases}.}
\label{fig:delta_influence_spatial}
\end{figure}

Figures~\ref{fig:kv_pathway_schematic} and~\ref{fig:kv_mechanism_cases} show the pathway and cases. For L21, K$+$V yields 3.5\% SR and mean $|\Delta A|=0.00390$; for L20, V-only yields 91.0\% SR and mean $|\Delta A|=4.75\times10^{-4}$.

\subsection{Predictive evidence from short probes}
\label{app:probe_prediction}

The shared component-ablation Probe100 Tail-AV scores correlate with later 5k/10k/20k AttackStrength at 0.344/0.521/0.508. Eight additional runs retain the same 30 candidates, task-balanced pairs, and 100-exposure budget; Table~\ref{tab:probe100_predictive_stability} reports predictive summaries and Top-$k$ overlaps.

\begin{figure}[!tbp]
    \centering
    \includegraphics[width=0.92\textwidth]{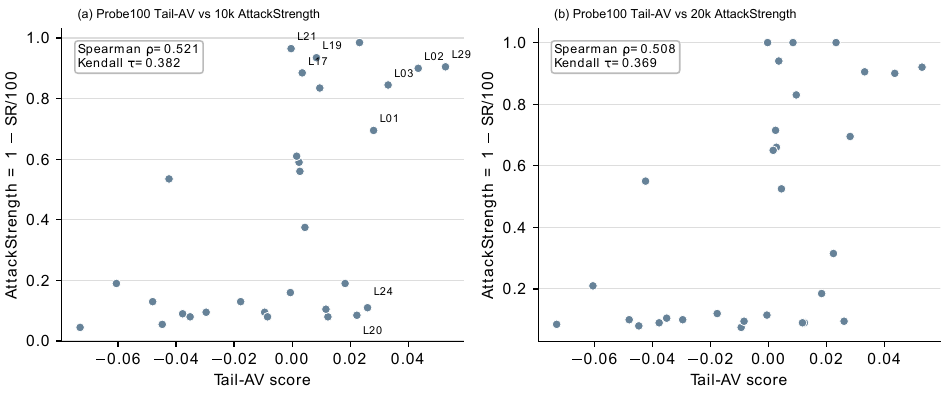}
\caption{\textbf{Probe100 Tail-AV scores versus later attack strength.} Each point is a FastWAM layer: Tail-AV from the shared 100-exposure component-ablation run versus later AttackStrength $=1-\mathrm{SR}/100$ from the matched BS=1 singleton patches at 10k or 20k exposures.}
    \label{fig:probe100_predictive}
\end{figure}

\begin{table}[H]
\centering
\caption{\textbf{Probe100 probe-seed stability across eight additional runs on the evaluated FastWAM checkpoint.} Each repeat uses the fixed scoring pairs and a complete 30-layer ranking; values report mean $\pm$ sample SD and observed range. The canonical ranking is used only as the Top-$k$ set reference.}
\label{tab:probe100_predictive_stability}
\footnotesize
\setlength{\tabcolsep}{4pt}
\begin{tabular}{@{}lcc@{}}
\toprule
Quantity & Mean $\pm$ SD & Observed range \\
\midrule
Spearman $\rho$ to 5k AttackStrength & $0.3550\pm0.0074$ & $[0.3465,0.3655]$ \\
Spearman $\rho$ to 10k AttackStrength & $0.5198\pm0.0080$ & $[0.5067,0.5297]$ \\
Spearman $\rho$ to 20k AttackStrength & $0.5065\pm0.0084$ & $[0.4953,0.5194]$ \\
Canonical Top-3 overlap & $0.9583\pm0.1179$ & $[0.6667,1.0000]$ \\
Canonical Top-5 overlap & $1.0000\pm0.0000$ & $[1.0000,1.0000]$ \\
\bottomrule
\end{tabular}
\end{table}

\FloatBarrier

\subsection{Complete layerwise mechanism summary}
\begingroup
\footnotesize
\setlength{\tabcolsep}{1.6pt}
\setlength{\LTcapwidth}{\textwidth}
\begin{longtable}{@{}lrrrrrrrrr@{}}
\caption{\textbf{Complete 30-layer K/V mechanism summary.} All rows use the joint K+V sweep with 50 observations from ten task-balanced tasks. $D_K$, $D_V$, and $D_A$ are representation/readout displacement summaries; AS is $1-\mathrm{SR}/100$. The decomposition and heatmaps provide per-observation detail.}\label{tab:fastwam_mechanism_full}\\
\toprule
Layer & SR (\%) & AS & $D_K$ & $D_V$ & $D_A$ & $\|\Delta_K\|$ & $\|\Delta_V\|$ & $\|\Delta_{KV}\|$ & Tail-AV \\
\midrule
\endfirsthead
\toprule
Layer & SR (\%) & AS & $D_K$ & $D_V$ & $D_A$ & $\|\Delta_K\|$ & $\|\Delta_V\|$ & $\|\Delta_{KV}\|$ & Tail-AV \\
\midrule
\endhead
\midrule
\endfoot
\bottomrule
\endlastfoot
L00 & 1.5 & 0.985 & 0.377 & 1.433 & 0.429 & 9.4 & 105.8 & 112.9 & 0.425 \\
L01 & 30.5 & 0.695 & 0.646 & 2.241 & 0.723 & 33.2 & 331.8 & 327.5 & 0.760 \\
L02 & 10.0 & 0.900 & 0.553 & 1.967 & 0.566 & 37.5 & 324.7 & 310.8 & 0.474 \\
L03 & 15.5 & 0.845 & 0.561 & 1.452 & 0.587 & 37.3 & 222.0 & 205.9 & 0.291 \\
L04 & 95.5 & 0.045 & 0.565 & 1.357 & 0.687 & 58.3 & 214.6 & 208.1 & 0.187 \\
L05 & 94.5 & 0.055 & 0.736 & 1.599 & 0.746 & 63.9 & 260.9 & 272.5 & 0.371 \\
L06 & 46.5 & 0.535 & 0.363 & 1.088 & 0.432 & 39.7 & 178.5 & 178.6 & -0.202 \\
L07 & 87.0 & 0.130 & 0.401 & 0.877 & 0.370 & 41.4 & 172.1 & 171.4 & -0.081 \\
L08 & 81.0 & 0.190 & 0.437 & 1.011 & 0.525 & 50.3 & 201.1 & 203.5 & -0.345 \\
L09 & 91.0 & 0.090 & 0.521 & 0.926 & 0.353 & 42.5 & 192.4 & 194.4 & -0.053 \\
L10 & 87.0 & 0.130 & 0.412 & 1.054 & 0.351 & 45.2 & 236.8 & 238.7 & -0.117 \\
L11 & 84.0 & 0.160 & 0.339 & 0.776 & 0.395 & 54.3 & 160.3 & 165.5 & -0.020 \\
L12 & 81.0 & 0.190 & 0.429 & 0.786 & 0.448 & 59.6 & 149.1 & 153.4 & 0.013 \\
L13 & 44.0 & 0.560 & 0.532 & 0.905 & 0.698 & 86.4 & 157.3 & 163.6 & -0.138 \\
L14 & 16.5 & 0.835 & 0.510 & 1.072 & 0.690 & 72.0 & 162.5 & 157.0 & -0.115 \\
L15 & 41.0 & 0.590 & 0.477 & 1.080 & 0.609 & 80.1 & 160.5 & 162.2 & -0.032 \\
L16 & 62.5 & 0.375 & 0.393 & 0.794 & 0.457 & 51.3 & 96.9 & 109.5 & 0.240 \\
L17 & 11.5 & 0.885 & 0.643 & 1.038 & 0.906 & 73.9 & 164.4 & 160.9 & 0.062 \\
L18 & 39.0 & 0.610 & 0.389 & 0.997 & 0.374 & 49.2 & 189.9 & 183.1 & -0.114 \\
L19 & 6.5 & 0.935 & 0.525 & 1.119 & 0.859 & 104.0 & 209.8 & 206.8 & -0.037 \\
L20 & 91.5 & 0.085 & 0.230 & 0.731 & 0.574 & 36.4 & 121.1 & 117.2 & 0.047 \\
L21 & 3.5 & 0.965 & 0.483 & 0.809 & 0.571 & 74.8 & 142.1 & 153.4 & 0.036 \\
L22 & 92.0 & 0.080 & 0.245 & 0.570 & 0.217 & 39.5 & 59.8 & 66.3 & -0.010 \\
L23 & 90.5 & 0.095 & 0.269 & 0.614 & 0.120 & 28.7 & 30.8 & 56.7 & -0.006 \\
L24 & 89.0 & 0.110 & 0.277 & 0.675 & 0.357 & 60.0 & 61.3 & 125.2 & 0.203 \\
L25 & 92.0 & 0.080 & 0.139 & 0.695 & 0.108 & 21.0 & 18.8 & 40.9 & 0.020 \\
L26 & 90.5 & 0.095 & 0.217 & 0.697 & 0.268 & 40.5 & 39.5 & 77.3 & 0.022 \\
L27 & 89.5 & 0.105 & 0.237 & 0.907 & 0.210 & 26.3 & 33.4 & 55.3 & 0.100 \\
L28 & 92.0 & 0.080 & 0.236 & 0.743 & 0.128 & 14.0 & 23.9 & 29.4 & 0.010 \\
L29 & 9.5 & 0.905 & 0.213 & 2.810 & 0.793 & 30.6 & 318.6 & 310.9 & 2.063 \\
\end{longtable}
\endgroup

\Needspace{300pt}
\subsection{Complete spatial attention and contribution maps}

\FloatBarrier
\section{Complete layerwise and PASA-KV results}
\label{app:layerwise}
\subsection{Architecture-wide component sweeps}
\label{app:component_sweeps}

This section reports pooled and complete layerwise component results.

\begin{table}[H]
    \caption{\textbf{All-layer component-objective attacks across three WAMs.} One evaluated checkpoint per architecture; values are pooled closed-loop Policy SR (\%; lower is stronger) over independently trained single-layer patches. K-only and V-only denote end-to-end patch optimization targets. Bold marks the lowest SR among matched-10k cells in each architecture.}
    \label{tab:component_attack_coverage_full}
    \centering
    \footnotesize
    \setlength{\tabcolsep}{3.8pt}
    \begin{tabular}{@{}lrrrr@{}}
        \toprule
        Architecture & Attn.-map & K only & V only & K+V \\
        \midrule
        FastWAM & 90.68 & 65.77 & 66.43 & \textbf{58.87} \\
        LingBot-VA & \textbf{92.45} & 93.37 & 92.63 & 93.18 \\
        Cosmos Policy & 91.34 & 89.91 & \textbf{89.79} & 90.04 \\
        \bottomrule
    \end{tabular}
    \vspace{2pt}

    \begin{minipage}{0.98\columnwidth}
    \footnotesize
    \textit{Protocol.} All objectives use BS1 and 10k exposures per independent patch. FastWAM/LingBot pool 30 layers $\times$ 200 episodes (6,000); Cosmos pools 28 $\times$ 200 (5,600), with K+V targeting current-primary slot 3 at joint=1. Objective definitions and full values: Appendix~\ref{app:component_sweeps}; separate 5k C/A/W/J-KV tests: Appendix~\ref{app:target_scope}.
    \end{minipage}
\end{table}

\paragraph{Attention-map objectives}
The attention-map columns are architecture-specific map displacement controls. FastWAM normalizes the action-to-first-frame spatial map and maximizes its clean--attacked center-of-mass distance, $\lVert\sum_j(a_j^p-a_j^0)r_j\rVert_2$, with zero action latents and video/action timesteps. LingBot uses first-action-step attention over renormalized current-frame agent-view columns; Cosmos uses action-slot-4 queries over current-primary slot 3. Their score is $\lVert A^p-A^0\rVert_F/\max(\lVert A^0\rVert_F,10^{-6})$. Each singleton objective uses detached clean targets.

\begingroup
\footnotesize
\setlength{\tabcolsep}{3.2pt}
\setlength{\LTcapwidth}{\textwidth}
\begin{longtable}{@{}lrrrr@{\hspace{12pt}}lrrrr@{}}
\caption{\textbf{Complete layerwise component attacks for the evaluated FastWAM and LingBot-VA checkpoints.} Each independently trained singleton uses BS1, 10k exposures and 200 evaluation episodes; values are Policy SR (\%). The FastWAM attention-map controls record training seed 20260801; other cells retain the formal protocol label 42 where reported.}\label{tab:component_layerwise_fast_ling}\\
\toprule
& \multicolumn{4}{c}{FastWAM} & & \multicolumn{4}{c}{LingBot-VA} \\
\cmidrule(lr){2-5}\cmidrule(lr){7-10}
Layer & Attn. & K & V & K+V & Layer & Attn. & K & V & K+V \\
\midrule
\endfirsthead
\toprule
& \multicolumn{4}{c}{FastWAM} & & \multicolumn{4}{c}{LingBot-VA} \\
\cmidrule(lr){2-5}\cmidrule(lr){7-10}
Layer & Attn. & K & V & K+V & Layer & Attn. & K & V & K+V \\
\midrule
\endhead
\midrule
\endfoot
\bottomrule
\endlastfoot
L00 & 89.5 & 83.5 & 31.5 & 1.5 & L00 & 94.0 & 96.0 & 93.0 & 91.5 \\
L01 & 87.5 & 31.5 & 28.0 & 30.5 & L01 & 90.0 & 93.5 & 90.0 & 89.5 \\
L02 & 91.0 & 9.5 & 10.0 & 10.0 & L02 & 95.0 & 93.0 & 92.0 & 95.0 \\
L03 & 91.5 & 87.5 & 6.0 & 15.5 & L03 & 90.0 & 94.0 & 95.0 & 94.0 \\
L04 & 91.5 & 67.5 & 91.0 & 95.5 & L04 & 93.0 & 94.0 & 90.5 & 93.0 \\
L05 & 93.0 & 94.0 & 89.5 & 94.5 & L05 & 89.0 & 95.5 & 89.5 & 91.5 \\
L06 & 90.5 & 79.0 & 46.5 & 46.5 & L06 & 90.5 & 95.0 & 92.0 & 94.0 \\
L07 & 92.0 & 89.5 & 92.5 & 87.0 & L07 & 92.0 & 91.5 & 95.0 & 94.0 \\
L08 & 90.5 & 73.0 & 94.0 & 81.0 & L08 & 93.0 & 93.5 & 92.5 & 95.0 \\
L09 & 92.0 & 91.0 & 91.5 & 91.0 & L09 & 94.5 & 95.0 & 93.5 & 93.5 \\
L10 & 88.0 & 88.0 & 89.5 & 87.0 & L10 & 91.5 & 95.5 & 90.5 & 94.0 \\
L11 & 92.0 & 80.5 & 82.0 & 84.0 & L11 & 94.0 & 93.0 & 95.0 & 92.5 \\
L12 & 91.5 & 85.5 & 88.0 & 81.0 & L12 & 91.0 & 91.0 & 92.0 & 92.0 \\
L13 & 89.5 & 80.0 & 25.5 & 44.0 & L13 & 92.0 & 88.0 & 92.0 & 92.5 \\
L14 & 91.0 & 68.0 & 22.0 & 16.5 & L14 & 94.0 & 91.0 & 90.5 & 94.5 \\
L15 & 87.0 & 64.0 & 39.0 & 41.0 & L15 & 91.5 & 92.0 & 89.0 & 93.5 \\
L16 & 93.0 & 68.5 & 87.5 & 62.5 & L16 & 92.5 & 92.0 & 91.5 & 93.5 \\
L17 & 90.0 & 66.0 & 91.5 & 11.5 & L17 & 94.0 & 94.0 & 93.0 & 93.5 \\
L18 & 90.5 & 7.0 & 38.0 & 39.0 & L18 & 92.5 & 92.0 & 92.5 & 92.5 \\
L19 & 94.0 & 9.5 & 10.5 & 6.5 & L19 & 93.5 & 95.0 & 94.0 & 94.0 \\
L20 & 89.0 & 0.0 & 91.0 & 91.5 & L20 & 91.0 & 94.5 & 92.0 & 91.0 \\
L21 & 93.0 & 6.0 & 91.0 & 3.5 & L21 & 90.5 & 91.0 & 94.0 & 95.0 \\
L22 & 89.0 & 88.5 & 93.0 & 92.0 & L22 & 94.0 & 92.0 & 92.0 & 93.0 \\
L23 & 94.0 & 10.5 & 91.5 & 90.5 & L23 & 92.0 & 97.5 & 94.5 & 92.0 \\
L24 & 87.0 & 91.0 & 90.5 & 89.0 & L24 & 93.5 & 96.5 & 95.5 & 94.0 \\
L25 & 88.0 & 90.0 & 95.0 & 92.0 & L25 & 92.5 & 91.0 & 92.5 & 92.5 \\
L26 & 90.0 & 90.5 & 88.0 & 90.5 & L26 & 92.0 & 93.5 & 91.5 & 94.5 \\
L27 & 91.5 & 91.0 & 92.5 & 89.5 & L27 & 93.5 & 93.5 & 96.0 & 94.0 \\
L28 & 92.5 & 90.0 & 92.0 & 92.0 & L28 & 93.5 & 91.5 & 93.0 & 94.0 \\
L29 & 91.0 & 92.5 & 14.5 & 9.5 & L29 & 93.5 & 95.5 & 95.0 & 92.0 \\
\midrule
Pooled & 90.68 & 65.77 & 66.43 & 58.87 & Pooled & 92.45 & 93.37 & 92.63 & 93.18 \\
\end{longtable}
\endgroup

\begingroup
\footnotesize
\setlength{\tabcolsep}{3.5pt}
\setlength{\LTcapwidth}{\textwidth}
\begin{longtable}{@{}lrrrrrrrr@{}}
\caption{\textbf{Complete layerwise component attacks for the evaluated Cosmos Policy checkpoint.} The first four columns are matched BS=1, formal 42-labeled, joint=1, 10k-exposure objectives; K+V targets current-primary slot 3. The last four columns report historical 5k C/A/W/J objectives. Every cell uses 200 episodes; Table~\ref{tab:component_attack_coverage_full} summarizes the matched 10k columns.}\label{tab:component_layerwise_cosmos}\\
\toprule
& \multicolumn{4}{c}{Matched 10k component objectives} & \multicolumn{4}{c}{Historical 5k target-scope objectives} \\
\cmidrule(lr){2-5}\cmidrule(lr){6-9}
Layer & Attn.-map & K only & V only & K+V & C-KV & A-KV & W-KV & J-KV \\
\midrule
\endfirsthead
\toprule
& \multicolumn{4}{c}{Matched 10k component objectives} & \multicolumn{4}{c}{Historical 5k target-scope objectives} \\
\cmidrule(lr){2-5}\cmidrule(lr){6-9}
Layer & Attn.-map & K only & V only & K+V & C-KV & A-KV & W-KV & J-KV \\
\midrule
\endhead
\midrule
\endfoot
\bottomrule
\endlastfoot
L00 & 89.5 & 92.0 & 92.0 & 93.0 & 92.0 & 90.5 & 92.5 & 91.5 \\
L01 & 92.5 & 89.0 & 92.0 & 91.0 & 90.5 & 90.5 & 94.0 & 93.0 \\
L02 & 95.0 & 88.5 & 88.0 & 88.5 & 89.5 & 91.0 & 91.5 & 89.5 \\
L03 & 91.0 & 91.0 & 91.0 & 88.5 & 90.5 & 91.0 & 90.0 & 90.0 \\
L04 & 90.5 & 90.0 & 89.5 & 89.0 & 87.5 & 92.5 & 90.5 & 91.0 \\
L05 & 92.5 & 89.5 & 89.5 & 89.0 & 88.5 & 92.0 & 90.5 & 92.5 \\
L06 & 93.5 & 90.0 & 89.5 & 90.5 & 90.0 & 87.0 & 87.0 & 88.5 \\
L07 & 89.5 & 90.0 & 90.0 & 90.5 & 90.5 & 90.5 & 88.0 & 91.0 \\
L08 & 90.0 & 91.5 & 91.0 & 91.5 & 90.5 & 91.0 & 89.5 & 92.5 \\
L09 & 89.0 & 90.5 & 92.5 & 92.0 & 90.5 & 87.5 & 89.5 & 90.0 \\
L10 & 87.5 & 91.5 & 91.0 & 91.0 & 90.5 & 91.5 & 91.0 & 88.5 \\
L11 & 92.5 & 90.5 & 89.0 & 89.0 & 90.5 & 92.5 & 90.0 & 92.0 \\
L12 & 92.0 & 89.0 & 90.0 & 91.0 & 90.0 & 91.5 & 92.5 & 92.0 \\
L13 & 91.0 & 89.0 & 91.0 & 90.5 & 90.0 & 90.0 & 88.0 & 87.5 \\
L14 & 94.0 & 90.5 & 91.5 & 90.0 & 89.5 & 92.0 & 93.0 & 90.5 \\
L15 & 94.5 & 90.5 & 89.5 & 90.0 & 91.5 & 93.5 & 92.0 & 90.5 \\
L16 & 92.5 & 87.0 & 87.0 & 87.5 & 90.5 & 90.0 & 91.0 & 86.5 \\
L17 & 91.5 & 88.0 & 88.0 & 88.5 & 90.0 & 91.0 & 91.0 & 89.5 \\
L18 & 89.5 & 90.0 & 90.0 & 90.0 & 89.5 & 88.0 & 89.0 & 89.5 \\
L19 & 92.0 & 90.5 & 89.0 & 90.0 & 88.0 & 92.5 & 88.5 & 87.0 \\
L20 & 91.5 & 89.0 & 89.0 & 90.5 & 89.0 & 87.5 & 88.5 & 90.5 \\
L21 & 92.0 & 89.0 & 88.5 & 87.5 & 88.5 & 88.0 & 89.0 & 89.5 \\
L22 & 88.5 & 90.0 & 93.0 & 92.5 & 89.0 & 92.0 & 88.5 & 91.5 \\
L23 & 88.0 & 89.0 & 89.5 & 89.0 & 91.5 & 90.5 & 88.0 & 89.0 \\
L24 & 95.0 & 91.0 & 87.5 & 87.0 & 90.0 & 92.0 & 91.5 & 89.0 \\
L25 & 93.5 & 91.5 & 86.0 & 91.0 & 90.0 & 90.0 & 91.5 & 87.5 \\
L26 & 89.0 & 90.5 & 89.5 & 91.5 & 90.5 & 90.5 & 90.0 & 89.0 \\
L27 & 90.0 & 89.0 & 90.0 & 91.0 & 91.0 & 93.5 & 89.5 & 93.5 \\
\midrule
Pooled & 91.34 & 89.91 & 89.79 & 90.04 & 89.98 & 90.71 & 90.20 & 90.09 \\
\end{longtable}
\endgroup

The matched 10k sweep has 352 rows across four objectives and 30/30/28 FastWAM, LingBot-VA, and Cosmos layers. The 112 historical 5k C/A/W/J rows supplement the architecture-specific sweep. Pooling sums successes and trials over layers; Table~\ref{tab:fastwam_kv_complementarity} reports paired statistics and the descriptive 50\% SR threshold.

\subsection{Exposure responses and selected multi-layer attacks}
\label{app:exposure_responses}

\paragraph{Post-hoc reporting groups}
The three-point trajectories define Early by $\mathrm{SR}_{10k}\leq35$ and $|\mathrm{SR}_{20k}-\mathrm{SR}_{10k}|\leq8$ percentage points; delayed-drop by $\mathrm{SR}_{5k},\mathrm{SR}_{10k}\geq80$, $\mathrm{SR}_{10k}-\mathrm{SR}_{20k}\geq15$, and $\mathrm{SR}_{20k}<80$; and stable-high by a three-point minimum of 78 and $|\mathrm{SR}_{5k}-\mathrm{SR}_{20k}|\leq12$. Remaining trajectories are Intermediate.

The display contains nine Early, five Intermediate (L06/L13/L15/L16/L18), and 16 Other targets; Other includes 15 stable-high trajectories and delayed-drop L20.

\begingroup
\footnotesize
\setlength{\tabcolsep}{3.2pt}
\setlength{\LTcapwidth}{\textwidth}
\begin{longtable}{@{}lrrrl@{\hspace{10pt}}lrrrl@{}}
\caption{Complete FastWAM singleton K/V exposure trajectories. Policy SR is in percent; labels are post-hoc reporting groups under the documented objective, exposure checkpoints, and rollout protocol.}\label{tab:layerwise_exposure}\\
\toprule
Layer & 5k & 10k & 20k & Response & Layer & 5k & 10k & 20k & Response \\
\midrule
\endfirsthead
\toprule
Layer & 5k & 10k & 20k & Response & Layer & 5k & 10k & 20k & Response \\
\midrule
\endhead
\midrule
\endfoot
\bottomrule
\endlastfoot
L00 & 64.5 & 1.5 & 0.0 & Early & L15 & 60.5 & 41.0 & 28.5 & Intermediate \\
L01 & 34.5 & 30.5 & 30.5 & Early & L16 & 92.0 & 62.5 & 47.5 & Intermediate \\
L02 & 10.0 & 10.0 & 10.0 & Early & L17 & 91.0 & 11.5 & 6.0 & Early \\
L03 & 91.0 & 15.5 & 9.5 & Early & L18 & 46.5 & 39.0 & 35.0 & Intermediate \\
L04 & 92.5 & 95.5 & 91.5 & Other & L19 & 38.5 & 6.5 & 0.0 & Early \\
L05 & 92.5 & 94.5 & 92.0 & Other & L20 & 90.0 & 91.5 & 68.5 & Other \\
L06 & 46.0 & 46.5 & 45.0 & Intermediate & L21 & 90.5 & 3.5 & 0.0 & Early \\
L07 & 89.0 & 87.0 & 90.0 & Other & L22 & 91.0 & 92.0 & 89.5 & Other \\
L08 & 87.5 & 81.0 & 79.0 & Other & L23 & 91.0 & 90.5 & 92.5 & Other \\
L09 & 92.0 & 91.0 & 91.0 & Other & L24 & 93.0 & 89.0 & 90.5 & Other \\
L10 & 89.5 & 87.0 & 88.0 & Other & L25 & 94.0 & 92.0 & 91.0 & Other \\
L11 & 78.5 & 84.0 & 88.5 & Other & L26 & 92.5 & 90.5 & 90.0 & Other \\
L12 & 86.0 & 81.0 & 81.5 & Other & L27 & 87.5 & 89.5 & 91.0 & Other \\
L13 & 60.0 & 44.0 & 34.0 & Intermediate & L28 & 91.0 & 92.0 & 90.5 & Other \\
L14 & 42.5 & 16.5 & 17.0 & Early & L29 & 40.0 & 9.5 & 8.0 & Early \\
\end{longtable}
\endgroup

\begin{figure}[!tbp]
\centering
\includegraphics[width=0.72\textwidth]{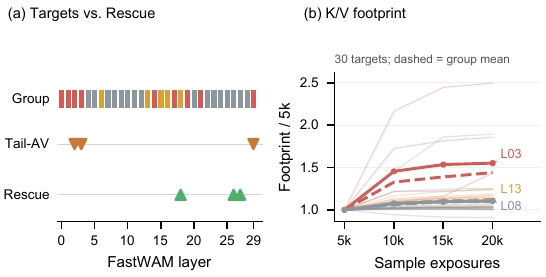}
\caption{\textbf{Layerwise intervention diagnostics on the evaluated FastWAM checkpoint.} \textbf{(a)} Tail-AV-selected targets and Rescue peaks. \textbf{(b)} K/V-footprint growth relative to each target's 5k value. Colors denote retrospective response groups. Detailed protocols and analyses are provided in Appendices~\ref{app:exposure_responses} and~\ref{app:layer_roles}.}
\label{fig:layer_mechanisms}
\end{figure}

\begin{samepage}
The Tail-AV-selected layer sets, listed in selected order for Top-3/4/5, are:
\begin{itemize}
\item \textbf{FastWAM}: L29/L02/L03, L29/L02/L03/L01, and L29/L02/L03/L01/L24.
\item \textbf{LingBot-VA}: L21/L02/L01, L21/L02/L01/L03, and L21/L02/L01/L03/L27.
\item \textbf{Cosmos Policy}: L14/L10/L01, L14/L10/L01/L26, and L14/L10/L01/L26/L19.
\end{itemize}
Each set is trained as a separate fresh joint patch.
\end{samepage}

\subsection{Footprint and visual-state replacement}
\label{app:layer_roles}

\paragraph{Independent all-target remeasurement}
Thirty targets use 5k/10k/15k/20k patches from one run per target (seed 20260801), giving $30\times4\times10\times30=36{,}000$ observed-layer records (1,200 paired conditions). BF16 forwards use tasks 0--9, frame 10, the documented frame-0 primary patch, and all 30 cache layers.

\paragraph{Footprint definition and aggregation}
For target $\ell$, training step $s$, task $t$, and observed layer $m$, let $d^K_{\ell stm}$ and $d^V_{\ell stm}$ be the clean-normalized tensor L2 displacements. Differences are evaluated in FP32 with clean norms clamped below at $10^{-12}$. We compute
\begin{equation}
F_\ell(s)=\sum_{m=0}^{29}\frac{1}{10}\sum_{t=0}^{9}\frac{d^K_{\ell stm}+d^V_{\ell stm}}{2},\qquad
R_\ell(s)=\frac{F_\ell(s)}{F_\ell(5\mathrm{k})}.
\label{eq:exact_footprint_all30}
\end{equation}
All 5k denominators are positive, and class means equally average target-normalized ratios. Tables~\ref{tab:exact_footprint} and~\ref{tab:footprint_class_summary} report footprint growth from the remeasurement runs, separately from historical Policy SR.

\begingroup
\footnotesize
\setlength{\tabcolsep}{7pt}
\setlength{\LTcapwidth}{\textwidth}
\begin{longtable}{@{}llrrrrr@{}}
\caption{\textbf{Independent all-target exact-checkpoint K/V footprints.} Absolute task-mean footprint $F_\ell(s)$ and target-normalized 20k/5k ratio $R_\ell(20\mathrm{k})$. All thirty targets use training seed 20260801, four checkpoints from the same run per target, ten paired tasks per point, and thirty observed layers.}\label{tab:exact_footprint}\\
\toprule
Target & Group & 5k & 10k & 15k & 20k & 20k/5k \\
\midrule
\endfirsthead
\toprule
Target & Group & 5k & 10k & 15k & 20k & 20k/5k \\
\midrule
\endhead
\midrule
\endfoot
\bottomrule
\endlastfoot
L00 & Early & 13.768 & 15.131 & 15.638 & 16.121 & 1.1709 \\
L01 & Early & 15.310 & 15.577 & 15.678 & 15.722 & 1.0269 \\
L02 & Early & 16.536 & 16.939 & 17.014 & 17.032 & 1.0300 \\
L03 & Early & 10.294 & 14.971 & 15.792 & 15.969 & 1.5513 \\
L04 & Other & 8.533 & 8.886 & 9.026 & 9.119 & 1.0687 \\
L05 & Other & 7.459 & 9.079 & 9.279 & 9.353 & 1.2539 \\
L06 & Intermediate & 13.619 & 13.996 & 14.155 & 14.197 & 1.0425 \\
L07 & Other & 7.427 & 7.616 & 7.652 & 7.676 & 1.0336 \\
L08 & Other & 10.386 & 11.161 & 11.363 & 11.479 & 1.1052 \\
L09 & Other & 7.693 & 7.954 & 8.046 & 8.112 & 1.0544 \\
L10 & Other & 8.723 & 9.536 & 9.737 & 9.836 & 1.1277 \\
L11 & Other & 8.842 & 8.324 & 8.084 & 8.034 & 0.9086 \\
L12 & Other & 7.834 & 8.044 & 8.057 & 7.811 & 0.9970 \\
L13 & Intermediate & 9.555 & 10.266 & 10.482 & 10.567 & 1.1060 \\
L14 & Early & 11.283 & 12.557 & 12.825 & 12.939 & 1.1468 \\
L15 & Intermediate & 10.011 & 11.663 & 12.176 & 12.431 & 1.2418 \\
L16 & Intermediate & 9.013 & 9.823 & 10.423 & 10.700 & 1.1872 \\
L17 & Early & 9.596 & 16.557 & 17.405 & 17.807 & 1.8557 \\
L18 & Intermediate & 7.536 & 7.699 & 7.757 & 7.782 & 1.0327 \\
L19 & Early & 15.196 & 17.167 & 17.645 & 21.792 & 1.4341 \\
L20 & Other & 7.645 & 11.105 & 14.222 & 14.471 & 1.8927 \\
L21 & Early & 7.069 & 15.291 & 17.277 & 17.648 & 2.4966 \\
L22 & Other & 6.736 & 6.777 & 6.752 & 6.762 & 1.0039 \\
L23 & Other & 6.623 & 6.745 & 6.730 & 6.679 & 1.0085 \\
L24 & Other & 6.371 & 6.415 & 6.417 & 6.417 & 1.0072 \\
L25 & Other & 6.243 & 6.318 & 6.348 & 6.367 & 1.0200 \\
L26 & Other & 6.291 & 6.398 & 6.487 & 6.531 & 1.0382 \\
L27 & Other & 6.101 & 6.201 & 6.246 & 6.273 & 1.0283 \\
L28 & Other & 6.291 & 6.321 & 6.355 & 6.364 & 1.0116 \\
L29 & Early & 13.876 & 16.866 & 16.952 & 17.193 & 1.2390 \\
\end{longtable}
\endgroup

\begin{table}[H]
\centering
\caption{\textbf{Class summaries of target-normalized footprints.} Class means average $R_\ell(s)$ after normalization within each target. Groups are retrospective; SD uses $n-1$ across target layers. All 5k ratios equal one.}
\label{tab:footprint_class_summary}
\footnotesize
\setlength{\tabcolsep}{4.5pt}
\begin{tabular}{@{}lrrrrrrl@{}}
\toprule
 & & \multicolumn{3}{c}{Class mean} & \multicolumn{3}{c}{20k across-layer distribution} \\
\cmidrule(lr){3-5}\cmidrule(l){6-8}
Group & $n$ & 10k & 15k & 20k & Median & SD & Range \\
\midrule
Early & 9 & 1.3269 & 1.3889 & 1.4390 & 1.2390 & 0.4795 & [1.0269, 2.4966] \\
Intermediate & 5 & 1.0758 & 1.1077 & 1.1220 & 1.1060 & 0.0910 & [1.0327, 1.2418] \\
Other & 16 & 1.0618 & 1.0937 & 1.0975 & 1.0309 & 0.2246 & [0.9086, 1.8927] \\
\bottomrule
\end{tabular}
\end{table}

\paragraph{Target coverage and descriptive groups}
Figure~\ref{fig:layer_mechanisms}(b) plots all targets and class means; L03/L13/L08 are closest to their group means. Mean 20k/5k ratios are 1.4390/1.1220/1.0975, spanning 0.9086--2.4966.

\begin{figure}[!tbp]
\centering
\includegraphics[width=0.92\textwidth]{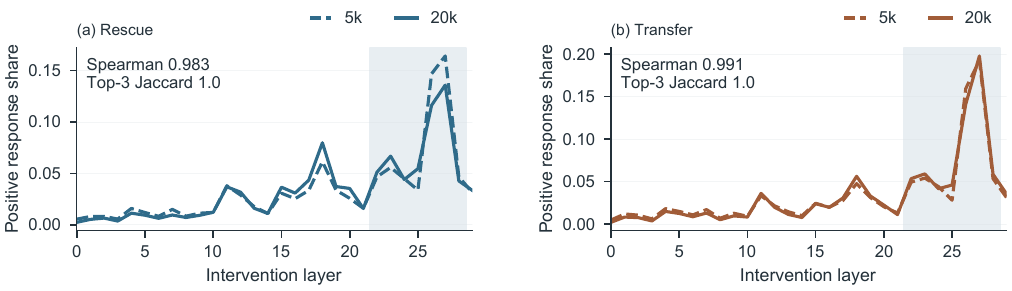}
\caption{\textbf{Visual-state replacement profiles.} Each layer averages 30 target patches and four fixed samples (tasks 0/1; frames 0/10); negative gains are clipped before across-layer normalization. Shading marks L22--L28. Five-k/20k profile correlations are 0.983/0.991 with unchanged Top-3 sets.}
\label{fig:replacement_profiles}
\end{figure}

\paragraph{Response definition and aggregation}
Rescue and Transfer measure normalized projected action change toward the clean and attacked action, respectively. Four fixed samples and then 30 targets are averaged to obtain $g_\ell$, and $q_\ell=\max(0,g_\ell)/\sum_m\max(0,g_m)$. Each budget has 7,200 offline intervention records; their reference Action-L2 rises from 0.7467 to 1.2499 across 5k/20k.

\paragraph{Online oracle pilot}
Table~\ref{tab:online_rescue} uses 10k singleton patches and 50 episodes per condition. An oracle substitutes visual K/V from paired clean/attacked caches at the same simulator state, then executes the mixed-cache action. L25 reuses weak-attack set A (L17--L19). The oracle requires paired clean observations at test time.

\begin{table}[H]
\centering
\caption{\textbf{Historical pilot of oracle K/V replacement.} Closed-loop Policy SR (\%; higher means better control here), 50 episodes per cell. These 10k singleton patches provide supplementary mechanism evidence.}
\label{tab:online_rescue}
\footnotesize
\setlength{\tabcolsep}{10pt}
\begin{tabular}{@{}lrrrrr@{}}
\toprule
Attack target & Clean & Attack & Set A (3 layers) & Set B (3 layers) & A $\cup$ B (6) \\
\midrule
L18 & 94 & 44 & 88 & 52 & 86 \\
L29 & 96 & 16 & 26 & 32 & 58 \\
L25 (weak control) & 94 & 88 & 88 & 94 & 98 \\
\bottomrule
\end{tabular}
\par\smallskip
\begin{minipage}{0.98\textwidth}
\footnotesize
At every replan, selected visual K/V states are replaced by a clean pass at the same simulator state. Set A is L17--L19 for L18/L25 and L27--L29 for L29; B is L24--L26. Each row uses its own paired task/trial grid (10 tasks $\times$ 5 trials).
\end{minipage}
\end{table}

\section{Baseline protocols and fidelity}
\label{app:baselines}

\paragraph{Rationale and interpretation}
Table~\ref{tab:baseline_fidelity} documents source-principle WAM adaptations, including geometry, training pools, and cost accounting; Figure~\ref{fig:selector_controls} isolates layer selection under a shared K/V objective and budget.

\begingroup
\footnotesize
\setlength{\tabcolsep}{4pt}
\renewcommand{\arraystretch}{1.15}
\setlength{\LTcapwidth}{\textwidth}
\begin{longtable}{@{}>{\raggedright\arraybackslash}p{0.15\linewidth}>{\raggedright\arraybackslash}p{0.20\linewidth}>{\raggedright\arraybackslash}p{0.27\linewidth}>{\raggedright\arraybackslash}p{0.30\linewidth}@{}}
\caption{\textbf{Baseline fidelity under our documented WAM adaptations.} Areas refer to the affected resized view, not the complete input.}\label{tab:baseline_fidelity}\\
\toprule
Method & Fidelity & Recorded geometry & Target / accessible representation \\
\midrule
\endfirsthead
\toprule
Method & Fidelity & Recorded geometry & Target / accessible representation \\
\midrule
\endhead
\midrule
\endfoot
\bottomrule
\endlastfoot
DP-Attacker & Source-principle adaptation &
FastWAM: $45\times45$, main, 4.04\%, EOT. Cosmos: $50\times50$, primary, 4.98\%. LingBot: $29\times29$, agent, $\sim5.13\%$. &
Action-denoising target / diffusion surrogate; no explicit K/V intervention. \\
\addlinespace
DRIFT & Source-principle adaptation; Cosmos EDM variant &
FastWAM: $32\times32$, wrist, 2.04\%, no EOT. Cosmos: $50\times50$, primary. LingBot: $18\times18$, wrist, $\sim1.98\%$, at $(55,110)$. &
Initial flow/velocity for FastWAM and LingBot; joint denoised prediction for Cosmos. \\
\addlinespace
UPA-RFAS & Conceptual adaptation; no PSM &
FastWAM/Cosmos: $50\times50$, primary. LingBot: $29\times29$, agent. &
Feature deviation, token contrastive repulsion, and action-query attention dominance; no explicit K/V intervention. \\
\addlinespace
UPA & Conceptual adaptation & Same geometry as UPA-RFAS. &
Continuous action-chunk objective. \\
\addlinespace
TMA & Conceptual adaptation & Same geometry as UPA-RFAS. &
Continuous action-chunk objective. \\
\addlinespace
UADA & Conceptual adaptation & Same geometry as UPA-RFAS. &
Continuous action-chunk objective. \\
\end{longtable}
\endgroup

\begin{table}[H]
\centering
\footnotesize
\setlength{\tabcolsep}{4pt}
\renewcommand{\arraystretch}{1.15}
\caption{Construction-cost accounting and observation-pool boundaries for the baseline adaptations.}
\label{tab:baseline_accounting}
\begin{tabular}{@{}>{\raggedright\arraybackslash}p{0.23\linewidth}>{\raggedright\arraybackslash}p{0.70\linewidth}@{}}
\toprule
Accounting field & Record and comparison boundary \\
\midrule
Updates / exposures & All six: BS=1, seed 42, 10,000 optimizer updates and 10,000 sample exposures. \\
GPU-hour basis & GPU-active or card-wall construction records, excluding victim pretraining, smoke tests, queueing, and rollout evaluation. LingBot DRIFT records training-loop time only (2.462206 GPU-h), also excluding loading. \\
Observation pool & DRIFT uses its historical Spatial pool, whereas DP-Attacker uses a task-balanced pool. \\
\bottomrule
\end{tabular}
\par\smallskip
\begin{minipage}{\linewidth}
\footnotesize
FastWAM UPA-RFAS/UPA/TMA/UADA form the closer-geometry subset relative to PASA-KV; DP and DRIFT use different geometries. Table~\ref{tab:main_attack_comparison} gives SR and recorded costs.
\end{minipage}
\end{table}

UPA-RFAS/UPA/TMA/UADA match PASA-KV's FastWAM geometry and retain 71.4--90.8\% SR. Costs are implementation-level workflow measurements (Section~\ref{app:costs}).

\paragraph{LingBot DRIFT noise pairing and evaluation}
During patch training, the clean and attacked velocity terms share noise per exposure. One-GPU BS1 training uses seed 42 and 10k updates (2.462206 GPU-h, excluding loading/evaluation); disjoint evaluation batches give $183/200+279/300=462/500$ successes.

\paragraph{UPA-RFAS contrastive reference set at batch size one}
The WAM translation uses token positions, rather than different batch samples, as its contrastive reference set. For one image with $P$ visual tokens, let $c_u$ and $a_v$ be unit-normalized clean (stop-gradient) and attacked features. The patch maximizes
\begin{equation}
J_{\mathrm{tok}}=-\frac{1}{P}\sum_{u=1}^{P}
\log\frac{\exp(c_u^\top a_u/\tau)}
{\sum_{v=1}^{P}\exp(c_u^\top a_v/\tau)}.
\end{equation}
Logits have shape $[B,P,P]$ with token-index labels, retaining $P$ denominator terms at BS1. This token-reference contrast differs from sample-level contrast and uses action rather than source-model text queries.

\subsection{Complete layerwise action-discrepancy measurements}
\begin{table}[H]
\centering
\footnotesize
\setlength{\tabcolsep}{3pt}
\caption{\textbf{Matched BS=1, 20k singleton Action-L2 and Policy SR across 30 FastWAM layers.} Each row is a separately trained single-layer joint K/V patch. Action-L2 is the mean unnormalized FP32 L2 norm of the full action-chunk difference over four fixed inputs (tasks 0/1, frames 0/10). Eval200 SR is matched by layer to Table~\ref{tab:layerwise_exposure}.}
\label{tab:fastwam_action_l2_all30}
\begin{tabular}{@{}lrr@{\hspace{1em}}lrr@{\hspace{1em}}lrr@{}}
\toprule
Layer & Action-L2 & SR (\%) & Layer & Action-L2 & SR (\%) & Layer & Action-L2 & SR (\%) \\
\midrule
L00 & 2.1042 & 0.0 & L10 & 0.9184 & 88.0 & L20 & 1.3762 & 68.5 \\
L01 & 2.1661 & 30.5 & L11 & 0.4562 & 88.5 & L21 & 2.5344 & 0.0 \\
L02 & 2.8538 & 10.0 & L12 & 0.3653 & 81.5 & L22 & 0.2111 & 89.5 \\
L03 & 2.9605 & 9.5 & L13 & 1.2415 & 34.0 & L23 & 0.2490 & 92.5 \\
L04 & 0.3385 & 91.5 & L14 & 1.9948 & 17.0 & L24 & 0.2514 & 90.5 \\
L05 & 0.8464 & 92.0 & L15 & 1.9247 & 28.5 & L25 & 0.1863 & 91.0 \\
L06 & 1.8924 & 45.0 & L16 & 1.0540 & 47.5 & L26 & 0.1964 & 90.0 \\
L07 & 0.3268 & 90.0 & L17 & 2.5732 & 6.0 & L27 & 0.2019 & 91.0 \\
L08 & 1.3173 & 79.0 & L18 & 0.7030 & 35.0 & L28 & 0.1749 & 90.5 \\
L09 & 0.2765 & 91.0 & L19 & 3.6645 & 0.0 & L29 & 2.1620 & 8.0 \\
\bottomrule
\end{tabular}
\end{table}

\section{Historical Cross-Architecture Target-Scope Sweeps}
\subsection{Historical target-scope stress tests}
\label{app:target_scope}

Table~\ref{tab:target_scope} reports complete singleton sweeps. LingBot uses 20k updates, a $29\times29$ current-frame patch, video endpoint $s=0.6$, and ten action steps; one-step current/current-plus-future3 and three-step current-only yield 5,584/5,608/5,768 successes of 6,000. Cosmos uses 5k updates, one joint step, and a current-primary $50\times50$ patch; conditioning/action-facing/future/joint K/V targets yield 5,039/5,080/5,051/5,045 successes of 5,600 over 28 layers.

\begin{table}[H]
\centering
\caption{\textbf{Historical target-scope stress tests.} Complete within-system singleton sweeps broaden the K/V objective while keeping RGB patch placement fixed; episode-pooled Policy SR (\%).}
\label{tab:target_scope}
\footnotesize
\begin{minipage}[t]{0.54\textwidth}
\centering
\textbf{LingBot-VA: 20k updates, 6,000 episodes/row}\\[2pt]
\setlength{\tabcolsep}{3pt}
\begin{tabular}{@{}lrrr@{}}
\toprule
K/V objective & Video steps & Clean & Attack SR \\
\midrule
Current & 1 & 93.50 & 93.07 \\
Current + future3 & 1 & 93.50 & 93.47 \\
Current & 3 & 95.50 & 96.13 \\
\bottomrule
\end{tabular}
\end{minipage}\hfill
\begin{minipage}[t]{0.44\textwidth}
\centering
\textbf{Cosmos: 5k updates, 5,600 episodes/row}\\[2pt]
\setlength{\tabcolsep}{5pt}
\begin{tabular}{@{}lrr@{}}
\toprule
Objective & Clean & Attack SR \\
\midrule
C-KV: conditioning & 97.00 & 89.98 \\
A-KV: action-facing & 97.00 & 90.71 \\
W-KV: future & 97.00 & 90.20 \\
J-KV: joint objective & 97.00 & 90.09 \\
\bottomrule
\end{tabular}
\end{minipage}
\par\smallskip
\begin{minipage}{0.98\textwidth}
\footnotesize
LingBot: 30 layers $\times$ 200 episodes, ten action steps; ``future3'' extends the K/V loss, not patch placement. Cosmos: one joint step, 28 layers $\times$ 200 episodes; distinct from the matched-10k component ablation. Native clean: 200 episodes. Full protocols: Appendix~\ref{app:target_scope}.
\end{minipage}
\end{table}

\end{document}